\documentclass{article} 
\usepackage{iclr2018_conference,times}
\usepackage{amsfonts}       
\usepackage{amssymb}
\usepackage{amsmath}
\usepackage{amsthm}
\usepackage{bm}
\usepackage{booktabs}       
\usepackage{color}
\usepackage{empheq}
\usepackage{graphicx}
\usepackage{hyperref}       
\usepackage{microtype}      
\usepackage{multirow}
\usepackage{nicefrac}       
\usepackage{url}            
\usepackage{wrapfig}

\title{Diffusion Convolutional Recurrent Neural Network: Data-Driven Traffic Forecasting}

\author{Yaguang Li$^\dag$, Rose Yu$^\ddag$, Cyrus Shahabi$^\dag$, Yan Liu$^\dag$ \\
$^\dag$ University of Southern California, $^\ddag$ California Institute of Technology \\
$^\dag$ \texttt{\{yaguang, shahabi, yanliu.cs\}@usc.edu}, $^\ddag$ \texttt{rose@caltech.edu} \\
}

\iclrfinalcopy

\begin{document}
\newcommand{\eat}[1]{}
\newcommand{\ryedit}[1]{\textcolor{blue}{\emph{[RY: #1]}}}
\newcommand{\cxbedit}[1]{\textcolor{cyan}{\emph{[CXB: #1]}}}
\newcommand{\tbedit}[1]{\textcolor{magenta}{\emph{[TB: #1]}}}
\newcommand{\dhedit}[1]{{\color{red} #1}}
\newcommand{\dhcomment}[1]{\dhedit{[DH: #1]}}
\newcommand{\yledit}[1]{{\color{blue} #1}}
\newcommand{\ylcomment}[1]{\yledit{[YL: #1]}}
\newcommand{\ygedit}[1]{{\color{blue} #1}}
\newcommand{\ygcomment}[1]{\ygedit{[YG: #1]}}
\newcommand{\todo}[1]{\textcolor{red}{\emph{[TODO: #1]}}}

\newenvironment{example}[1][Example]{\begin{trivlist}
\item[\hskip \labelsep {\bfseries #1}]}{\end{trivlist}}
\newenvironment{remark}[1][Remark]{\begin{trivlist}
\item[\hskip \labelsep {\bfseries #1}]}{\end{trivlist}}

\newtheorem{definition}{Definition}
\newtheorem{problem}{Problem}
\newtheorem{theorem}{Theorem}[section]
\newtheorem{lemma}[theorem]{Lemma}
\newtheorem{note}[theorem]{Note}
\newtheorem{corollary}[theorem]{Corollary}
\newtheorem{prop}[theorem]{Proposition}

\newcommand{\abs}[1]{ {\left| #1 \right|}}
\newcommand{\pq}[1]{\left( #1 \right)}
\newcommand{\cpd}[1]{{\left\llbracket {#1} \right\rrbracket}}

\newcommand{\mb}[1]{{\mathbf{#1}}}
\newcommand{\V}[1]{{\bm{#1}}} 
\newcommand{\M}[1]{{\bm{#1}}} 
\newcommand{\T}[1]{{\mb{\mathsf{#1}}}} 
\newcommand{\C}[1]{\mathcal{#1}} 

\newcommand{\adj}{\M{A}}
\newcommand{\gweights}{{\M{W}}}
\newcommand{\ex}{\mathbb{E}}
\newcommand{\edges}{\C{E}}
\newcommand{\graph}{\C{G}}
\newcommand{\vertices}{\C{V}}
\newcommand{\real}{\mathbb{R}}
\newcommand{\skt}{{\mathcal{S}}}
\newcommand{\lapl}{\M{L}}
\newcommand{\loss} {\mathcal{L}}
\newcommand{\ppr}{{\C{P}}}  

\newcommand{\hprod}{\odot}
\newcommand{\kprod}{\otimes}
\newcommand{\krprod}{\odot}
\newcommand{\oprod}{\circ }
\newcommand{\gconv}{{\,\star_{\mathcal{G}}\,}}
\newcommand{\spectral}[1]{\widehat{#1}}

\newcommand{\SM}[1]{\mathbf{\hat{#1}}}
\newcommand{\transpose}{\intercal}
\newcommand{\bestval}[1]{{\textbf{#1}}}

\newcommand{\SV}[2]{\sigma_{#1}(#2)}
\newcommand{\EV}[2]{\lambda_{#1}(#2)}
\newcommand{\numberthis}{\addtocounter{equation}{1}\tag{\theequation}}

\newcommand{\gcrnn}{{DCRNN}}

\newcommand{\dummyfigure}[2]{\fbox{\begin{minipage}{#1}\hfill\vspace{#2}\end{minipage}}}

\maketitle
 \vspace{-0.2in}
\begin{abstract}
 \vspace{-0.1in}
Spatiotemporal forecasting has various applications in neuroscience, climate and transportation domain. 
Traffic forecasting is one canonical example of such learning task.
The task is challenging  due to  (1) complex spatial dependency on road networks, (2) non-linear temporal dynamics  with changing road conditions and (3) inherent difficulty of long-term forecasting. 
To address these challenges, we propose to model the traffic flow as a diffusion process on a directed graph and introduce \textit{Diffusion Convolutional Recurrent Neural Network} (\gcrnn{}), a deep learning framework for traffic forecasting that incorporates both spatial and temporal dependency in the traffic flow. 
Specifically,  \gcrnn{} captures the spatial dependency using bidirectional random walks on the graph, and the temporal dependency using the encoder-decoder architecture with scheduled sampling. We evaluate the framework on two real-world large scale road network traffic datasets and observe consistent improvement of $12 \%$ - $15 \%$ over state-of-the-art baselines. 
\end{abstract}

\section{Introduction}

\vspace{-0.1in}
Spatiotemporal forecasting is a crucial  task for a learning system that operates in a dynamic environment. It has a wide range of applications  from autonomous vehicles operations, to energy and smart grid optimization, to logistics and supply chain management.  In this paper, we study one important task: traffic forecasting on road networks, the core component of the intelligent transportation systems. The goal of traffic forecasting is to predict the future traffic speeds of a sensor network given historic traffic speeds and the underlying road networks. 

\begin{wrapfigure}{R}{0.5\textwidth}
	\centering
    \includegraphics[width=\linewidth]{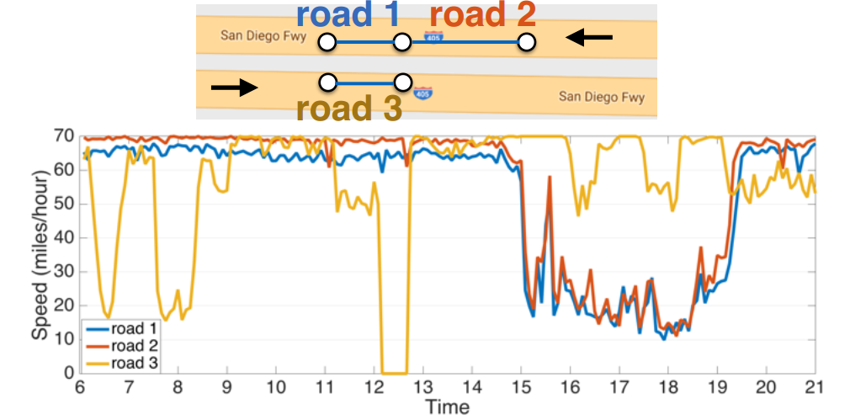}
    \caption{Spatial correlation is dominated by road network structure. (1) Traffic speed in road 1 are similar to road 2 as they locate in the same highway. (2) Road 1 and road 3 locate in the opposite directions of the highway. Though close to each other in the Euclidean space, their road network distance is large, and their traffic speeds differ significantly.}
    \label{fig:spatial}
\end{wrapfigure}

This task is challenging mainly due to the complex spatiotemporal dependencies and inherent difficulty in the long term forecasting. On the one hand, traffic time series demonstrate strong \textit{temporal dynamics}. Recurring incidents such as rush hours or accidents can cause non-stationarity, making it difficult to forecast long-term. On the other hand, sensors on the road network contain complex yet unique \textit{spatial correlations}.  Figure~\ref{fig:spatial} illustrates an example.  Road $1$ and road $2$ are correlated, while road $1$ and road $3$ are not. Although road $1$ and road $3$ are  close in the Euclidean space, they demonstrate very different behaviors.  Moreover, the future traffic speed is  influenced more by the downstream traffic than the upstream one. This means that the spatial structure in traffic is non-Euclidean and directional.

Traffic forecasting has been studied for decades, falling into two main categories:  knowledge-driven approach and data-driven approach. In transportation and operational research, knowledge-driven methods usually apply queuing theory and simulate user behaviors in traffic~\citep{cascetta2013transportation}. In time series community, data-driven methods such as Auto-Regressive Integrated Moving Average (ARIMA) model and Kalman filtering  remain popular~\citep{liu2011discovering, lippi2013short}. 
However, simple time series models usually rely on the stationarity assumption, which is often violated by the traffic data.  Most recently, deep learning models for traffic forecasting have been developed in~\citet{lv2015traffic, deeptraffic2017}, but without considering the spatial structure. \citet{wu2016short} and~\citet{ma2017learning} model the spatial correlation with Convolutional Neural Networks (CNN), but the spatial structure is in the Euclidean space (e.g., 2D images).   ~\citet{bruna2013spectral}, \citet{defferrard2016convolutional} studied graph convolution, but only for undirected graphs.  

In this work, we represent the pair-wise spatial correlations between traffic sensors using a directed graph whose nodes are sensors and edge weights denote proximity between the sensor pairs measured by the road network distance. We model the dynamics of the traffic flow as a diffusion process and propose the \emph{diffusion convolution} operation to capture the spatial dependency. 
We further propose \emph{Diffusion Convolutional Recurrent Neural Network} (\gcrnn{}) that integrates  \textit{diffusion convolution}, the \textit{sequence to sequence} architecture and  the \textit{scheduled sampling} technique.
When evaluated on real-world traffic datasets, \gcrnn{} consistently outperforms state-of-the-art traffic forecasting baselines by a large margin.
In summary:
\begin{itemize}
    \item We study the traffic forecasting problem and model the spatial dependency of traffic as a diffusion process on a directed graph. We propose \textit{diffusion convolution}, which has an intuitive interpretation and can be computed efficiently. 
    \item We propose \emph{Diffusion Convolutional Recurrent Neural Network} (\gcrnn{}), a holistic approach that captures both spatial and temporal dependencies among time series using \emph{diffusion convolution} and the sequence to sequence learning framework together with scheduled sampling. \gcrnn{} is not limited to transportation and is readily applicable to other spatiotemporal forecasting tasks. 
    \item We conducted extensive experiments on two large-scale real-world datasets, and the proposed approach obtains significant improvement over state-of-the-art baseline methods. 
\end{itemize}
\vspace{-0.1in}


\vspace{-0.1in}
\section {Methodology}
\vspace{-0.1in}
We formalize the learning problem of spatiotemporal traffic forecasting and describe how to model  the dependency structures using  \emph{diffusion convolutional recurrent neural network}. 
\vspace{-0.1in}
\subsection{Traffic Forecasting Problem}
\vspace{-0.1in}
The goal of traffic forecasting is to predict the future traffic speed given previously observed traffic flow from $N$  correlated sensors on the road network.  
We can represent the  sensor network as a weighted directed graph
$\graph = (\vertices, \edges, \gweights)$, where $\vertices$ is a set of nodes $\abs{\vertices} = N$, $\edges$ is a set of edges and $\gweights{} \in \real^{N \times N}$ is a weighted adjacency matrix representing the nodes proximity (e.g., a function of their road network distance). 
Denote the traffic flow observed on $\graph$ as a graph signal $\M{X} \in \real^{N\times P}$, where $P$ is the number of features of each node (e.g., velocity, volume). Let $\M{X}^{(t)}$ represent the graph signal observed at time $t$, the traffic forecasting problem aims to learn a function $h(\cdot)$ that maps $T'$ historical graph signals to future $T$ graph signals, given a graph  $\graph$: 
\[   [ \M{X}^{(t-T'+1)},  \cdots, \M{X}^{(t)}; \graph ] \xrightarrow[]{h(\cdot)} [\M{X}^{(t+1)}, \cdots, \M{X}^{(t+T)}] \]

\subsection{Spatial Dependency Modeling}
\label{sec:spatial_dependency}

We model the spatial dependency by relating traffic flow to a diffusion process, which explicitly captures  the stochastic nature of  traffic  dynamics.  This diffusion process is characterized by a random walk on $\graph$ with restart probability $\alpha \in [0, 1]$, and a state transition matrix $\M{D}_O^{-1} \M{W}$.  Here $\M{D_O} = \mathrm{diag}(\M{W} \V{1})$ is the out-degree diagonal matrix, and $\V{1}\in \real^N$ denotes the all one vector. 
After many time steps, such Markov process converges to  a stationary distribution 
$\M{\ppr} \in \real^{N\times N}$  whose $i$th row $\V{\ppr}_{i, :} \in \real{}^N$  represents the likelihood of diffusion from node $v_i \in \vertices$, hence the proximity w.r.t. the node $v_i$.  The following Lemma provides a closed form solution for the stationary distribution.
\begin{lemma}~\citep{teng2016scalable}
The stationary distribution of the diffusion process can be represented as a weighted combination of infinite random walks on the graph, and be calculated in closed form:
\begin{equation}
\M{\ppr} =  \sum_{k=0}^\infty  \alpha(1-\alpha) ^k \left(\M{D}_O^{-1} \M{W} \right)^k
\label{eq:ppr}
\end{equation}
\end{lemma}
where $k$ is the diffusion step. In practice,  we use a finite $K$-step truncation of the diffusion process and  assign a trainable weight to each step. 
%
We also include the reversed direction diffusion process, such that  the bidirectional diffusion offers the model more flexibility to capture the influence from both the upstream and the downstream traffic. 
\paragraph{Diffusion Convolution}
The resulted diffusion convolution operation over a graph signal $\M{X}  \in \real^{N \times P}$ and a filter $f_\M{\theta}$  is defined as:
\begin{equation}
\M{X}_{:, p} \gconv{} f_\M{\theta} = \sum_{k=0}^{K-1} \left(\theta_{k,1} \left(\M{D}_O^{-1} \M{W}\right)^k + \theta_{k, 2} \left(\M{D}_I^{-1}\M{W}^\intercal\right)^k \right) \M{X}_{:, p} \quad \text{for } p \in \{1, \cdots, P\} 
\label{eq:gconv}
\end{equation}
where $\M{\theta} \in \real^{K\times 2}$ are the parameters for the filter and  $\M{D}_O^{-1} \M{W}, \M{D}_I^{-1}\M{W}^\intercal$ represent the transition matrices of the diffusion process and the reverse one, respectively.
In general, computing the convolution can be expensive. However, if $\graph$ is sparse, Equation~\ref{eq:gconv} can be calculated efficiently using $O(K)$ recursive sparse-dense matrix multiplication with total time complexity $O(K|\edges|) \ll O(N^2)$. 
See Appendix~\ref{sec:recursive_calculation} for more detail.

\paragraph{Diffusion Convolutional Layer}
With the convolution operation defined in Equation~\ref{eq:gconv}, we can build a  diffusion convolutional layer that maps $P$-dimensional features to $Q$-dimensional outputs.  Denote the parameter tensor  as  $\M{\T{\Theta}} \in \real^{Q \times P \times K \times 2} = [\M{\theta}]_{q,p}$, where $\M{{\Theta}}_{q, p, :, :} \in \real^{K \times 2}$ parameterizes the convolutional filter for the $p$th input and the $q$th output. The diffusion convolutional layer is thus:
\begin{equation}
	\M{H}_{:, q} = \V{a} \left(\sum_{p=1}^{P} \M{X}_{:, p} \gconv{} f_{\M{\T{\Theta}}_{q, p, :, :}} \right)  \qquad \text{for} \; q \in \{1, \cdots, Q\}
    \label{eq:gconv_layer}
\end{equation}
where $\M{X} \in \real^{N \times P}$ is the input,  $\M{H} \in \real^{N \times Q}$ is the output,  $\{ f_{\M{\T{\Theta}}_{q, p, , :}} \}$ are the filters and  $\V{a}$ is the activation function (e.g., ReLU, Sigmoid).
Diffusion convolutional layer learns the representations for graph structured data and we can train it  using stochastic gradient based method.

\paragraph{Relation with Spectral Graph Convolution}

Diffusion convolution is defined on both directed and undirected graphs.  When applied to undirected graphs, we show that many existing graph structured convolutional operations including the popular spectral graph convolution, i.e., ChebNet~\citep{defferrard2016convolutional}, can be considered as a special case of diffusion convolution (up to a similarity transformation).
Let $\M{D}$ denote the degree matrix, and  $\M{L}=\M{D}^{-\frac{1}{2}} (\M{D} - \M{W})\M{D}^{-\frac{1}{2}} $ be the normalized graph Laplacian, the following Proposition demonstrates the connection.
\begin{prop}
The spectral graph convolution defined as  
\[\M{X}_{:,p} \gconv{} f_\V{\theta} =  \M{\Phi} \;F(\V{\theta}) \;\M{\Phi}^\intercal \M{X}_{:, p} \] 
with  eigenvalue decomposition $\M{L}=\mathbf{\Phi} \Lambda \mathbf{\Phi}^\intercal$ and  $F(\V{\theta}) = \sum_{0}^{K-1} \theta_k \M{\Lambda}^k$, is equivalent to graph diffusion convolution up to a similarity transformation, when the graph $\graph$ is undirected. 
\label{eq:spectral_conv}
\end{prop} 
\proof See Appendix~\ref{sec:gconv_relationship}.

\begin{figure}[tbp]
    \centering
    \includegraphics[width=.9\linewidth]{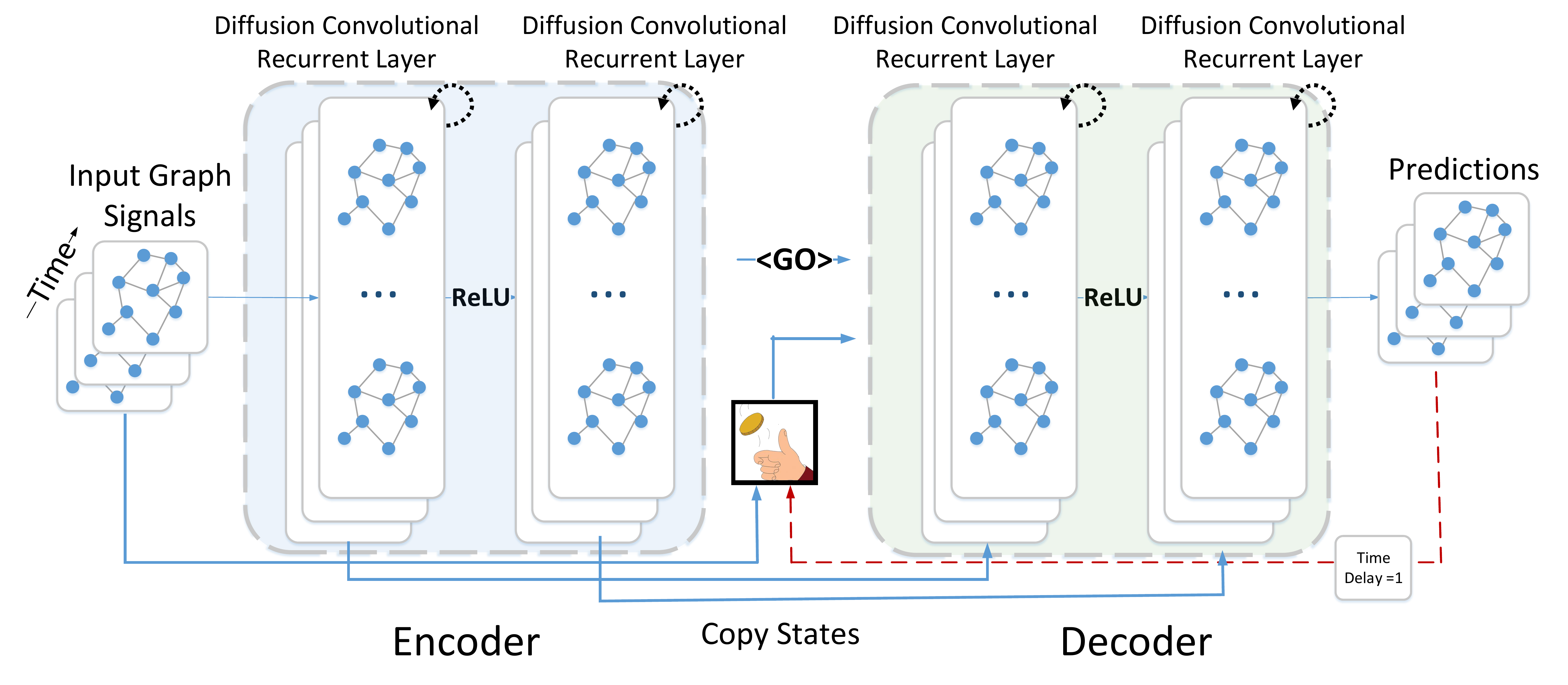}
    \caption{System architecture for the \textit{Diffusion Convolutional Recurrent Neural Network} designed for spatiotemporal traffic forecasting. The historical time series are fed into an encoder whose final states are used to initialize the decoder. The decoder makes predictions based on either previous ground truth or the model output.}
    \label{fig:system_architecture}
\end{figure}

\vspace{-0.1in}
\subsection{Temporal Dynamics Modeling}
\vspace{-0.1in}

We leverage the recurrent neural networks (RNNs)  to model the temporal dependency. In particular, we use Gated Recurrent Units (GRU)~\citep{chung2014empirical}, which is a simple yet powerful variant of RNNs.  We replace the matrix multiplications in GRU with the \textit{diffusion convolution}, which leads to our proposed \emph{Diffusion Convolutional Gated Recurrent Unit} (DCGRU).
\begin{eqnarray*}
\V{r}^{(t)} =& \sigma (\M{\Theta}_r \gconv{} [\M{X}^{(t)},\; \M{H}^{(t-1)}] + \V{b}_r)  \;
&\V{u}^{(t)} = \sigma (\M{\Theta}_u \gconv{} [\M{X}^{(t)},\; \M{H}^{(t-1)}] + \V{b}_u)   \\
\M{C}^{(t)} =& \tanh(\M{\Theta}_C \gconv{} \left[\M{X}^{(t)},\;  (\V{r}^{(t)} \hprod \M{H}^{(t-1)})\right] + \V{b}_c)  \;
&\M{H}^{(t)} = \V{u}^{(t)} \hprod{} \M{H}^{(t-1)} + (1-\V{u}^{(t)})\hprod{} \M{C}^{(t)}
\end{eqnarray*}
where $\M{X}^{(t)}, \M{H}^{(t)}$ denote the input and output of at time $t$, 
$\V{r}^{(t)}, \V{u}^{(t)}$ are reset gate and update gate at time $t$, respectively.
$\gconv{}$ denotes the \textit{diffusion convolution} defined in Equation~\ref{eq:gconv} and $\M{\Theta}_{r}, \M{\Theta}_{u}, \M{\Theta}_{C}$ are parameters for the corresponding filters.
Similar to GRU, DCGRU can be used to build recurrent neural network layers and be trained using backpropagation through time.

In multiple step ahead forecasting, we employ the \emph{Sequence to Sequence} architecture~\citep{sutskever2014sequence}. Both the encoder and the decoder are recurrent neural networks with DCGRU. 
During training, we feed the historical time series into the encoder and use its final states to initialize the decoder.  The decoder generates predictions given previous \emph{ground truth observations}.
At testing time, ground truth observations are replaced by predictions generated by the model itself.
The discrepancy between the input distributions of training and testing can cause degraded performance.
To mitigate this issue, we integrate \textit{scheduled sampling}~\citep{bengio2015scheduled} into the model, where we feed the model with either the ground truth observation with probability $\epsilon_i$ or the prediction by the model with probability $1-\epsilon_i$ at the $i$th iteration. During the training process, $\epsilon_i$ gradually decreases to $0$ to allow the model to learn the testing distribution.

With both spatial and temporal modeling, we build a \emph{Diffusion Convolutional Recurrent Neural Network}  (\gcrnn{}). The model  architecture of \gcrnn{} is shown in Figure ~\ref{fig:system_architecture}. The entire network is trained by maximizing the likelihood of generating the target future time series using backpropagation through time. \gcrnn{} is able to capture spatiotemporal dependencies among time series and can be applied to various spatiotemporal forecasting problems.

\vspace{-0.1in}
\section{Related Work}
\vspace{-0.1in}
Traffic forecasting is a classic problem in transportation and operational research which are primarily based on queuing theory and simulations~\citep{drew1968traffic}. 
Data-driven approaches for traffic forecasting have received considerable attention, and more details can be found in a recent survey paper~\citep{vlahogianni2014short} and the references therein.
However, existing machine learning models either impose strong stationary assumptions on the data (e.g., auto-regressive model) or fail to account for highly non-linear temporal dependency (e.g., latent space model~\citet{yu2016temporal, DengSDZYL16}).
Deep learning models deliver new promise for time series forecasting problem. 
For example, in~\citet{deeptraffic2017, laptev2017time}, the authors study time series forecasting using deep Recurrent Neural Networks (RNN). 
Convolutional Neural Networks (CNN) have also been applied  to traffic forecasting.
\citet{zhang2016dnn, zhang2017deep} convert the road network to a regular 2-D grid and apply traditional CNN to predict crowd flow. 
\cite{cheng2017deeptransport} propose DeepTransport which models the spatial dependency by explicitly collecting upstream and downstream neighborhood roads for each individual road and then conduct convolution on these neighborhoods respectively.

Recently, CNN has been generalized to arbitrary graphs based on the spectral graph theory. Graph convolutional neural networks (GCN) are first introduced in ~\citet{bruna2013spectral}, which bridges the spectral graph theory and deep neural networks. \citet{defferrard2016convolutional} propose ChebNet which improves GCN with fast localized convolutions filters. 
\citet{kipf2016semi} simplify ChebNet and achieve state-of-the-art performance in semi-supervised classification tasks.
\citet{seo2016structured} combine ChebNet with Recurrent Neural Networks (RNN) for structured sequence modeling.
~\citet{yu2017spatio} model the sensor network as a undirected graph and applied ChebNet and convolutional sequence model~\citep{gehring2017convolutional} to do forecasting.
One limitation of the mentioned spectral based convolutions is that they generally require the graph to be undirected to calculate meaningful spectral decomposition. 
Going from spectral domain to vertex domain,~\citet{atwood2016diffusion} propose diffusion-convolutional neural network (DCNN) which defines convolution as a diffusion process across each node in a graph-structured input. 
\cite{hechtlinger2017generalization} propose GraphCNN to generalize convolution to graph by convolving every node with its $p$ nearest neighbors.
However, both these methods do not consider the temporal dynamics and mainly deal with static graph settings. 

Our approach is different from all those methods due to both the problem settings and the formulation of the convolution on the graph. 
We model the sensor network as a weighted directed graph which is more realistic than grid or undirected graph.
Besides, the proposed convolution is defined using bidirectional graph random walk and is further integrated with the sequence to sequence learning framework as well as the scheduled sampling to model the long-term temporal dependency.

\vspace{-0.1in}
\section{Experiments}
\vspace{-0.1in}
We conduct experiments on two real-world large-scale datasets: (1) \textbf{METR-LA} This traffic dataset contains traffic information collected from loop detectors in the highway of Los Angeles County~\citep{Jagadish2014}.  We select 207 sensors and collect 4 months of data ranging from Mar 1st 2012 to Jun 30th 2012 for the experiment.
(2) \textbf{PEMS-BAY} This traffic dataset is collected by California Transportation Agencies (CalTrans) Performance Measurement System (PeMS). 
We select 325 sensors in the Bay Area and collect 6 months of data ranging from Jan 1st 2017 to May 31th 2017 for the experiment.
The sensor distributions of both datasets are visualized in~ Figure~\ref{fig:sensor_distribution} in the Appendix. 

In both of those datasets, we aggregate traffic speed readings into 5 minutes windows, and apply Z-Score normalization. 70\% of data is used for training, 20\% are used for testing while the remaining 10\% for validation. 
To construct the sensor graph, we compute the pairwise road network distances between sensors and build the adjacency matrix using thresholded Gaussian kernel~\citep{shuman2013emerging}. 
$
W_{ij} = \exp{\left(-\frac{\mathrm{dist}({v_i, v_j})^2}{\sigma^2}\right)} \quad \text{if } \mathrm{dist}(v_i, v_j) \leq \kappa \text{, otherwise } 0,
$
where $W_{ij}$ represents the edge weight between sensor $v_i$ and sensor $v_j$, $\mathrm{dist}({v_i, v_j})$ denotes the road network distance from sensor $v_i$ to sensor $v_j$. $\sigma$ is the standard deviation of distances and $\kappa$ is the threshold.
\vspace{-0.15in}
\subsection{Experimental Settings}
\vspace{-0.1in}
\paragraph{Baselines}

\begin{table}[tbp]
\centering
\caption{Performance comparison of different approaches for traffic speed forecasting. \gcrnn{} achieves the best performance with all three metrics for all forecasting horizons, and the advantage becomes more evident with the increase of the forecasting horizon.}
\label{tab:la_comparison}
\resizebox{\columnwidth}{!}{
\begin{tabular}{c||c|c|cccccccc}

\toprule
& $T$ & Metric & HA     & ARIMA$_{Kal}$  & VAR    & SVR    & FNN    & FC-LSTM  & \emph{\gcrnn{}}  \\ \hline
\midrule
\multirow{9}{*}{\rotatebox[origin=c]{90}{METR-LA}}&\multirow{3}{*}{15 min}& MAE  & 4.16       & 3.99   & 4.42   & 3.99   &  3.99  &  3.44 &  \bestval{2.77}  \\ 
& & RMSE & 7.80      & 8.21  & 7.89   & 8.45   &  7.94   & 6.30  &  \bestval{5.38}   \\ 
& & MAPE & 13.0\%  & 9.6\% & 10.2\% & 9.3\% &9.9\% & 9.6\%  & \bestval{7.3}\% \\ 
\cline{2-10}
& \multirow{3}{*}{30 min}& MAE  & 4.16     & 5.15   & 5.41   & 5.05   &4.23& 3.77 & \bestval{3.15}  \\ 
& & RMSE & 7.80     & 10.45   & 9.13   & 10.87   & 8.17   & 7.23    & \bestval{6.45}   \\ 
& & MAPE & 13.0\%  & 12.7\% & 12.7\% & 12.1\% & 12.9\% & 10.9\% &  \bestval{8.8}\% \\ 
\cline{2-10}
& \multirow{3}{*}{1 hour}& MAE  & 4.16    &  6.90  &  6.52 &  6.72  &   4.49 &  4.37 & \bestval{3.60}   \\ 
& & RMSE & 7.80     &  13.23  & 10.11  &  13.76  &  8.69  &  8.69  &  \bestval{7.59}   \\ 
& & MAPE & 13.0\%  &  17.4\% & 15.8\% & 16.7\%   &  14.0\%  &  13.2\% &   \bestval{10.5\%}   \\ 
\midrule
\midrule
\multirow{9}{*}{\rotatebox[origin=c]{90}{PEMS-BAY}}&\multirow{3}{*}{15 min}
& MAE  & 2.88   & 1.62   & 1.74   & 1.85   &  2.20  &  2.05 &  \bestval{1.38}  \\ 
& & RMSE & 5.59 & 3.30  & 3.16   &  3.59  &   4.42  &  4.19 &  \bestval{2.95}   \\ 
& & MAPE & 6.8\%  & 3.5\% & 3.6\% & 3.8\% & 5.19\% & 4.8\%  & \bestval{2.9}\% \\ 
\cline{2-10}
& \multirow{3}{*}{30 min}
  & MAE  & 2.88     & 2.33   & 2.32   & 2.48  &  2.30 & 2.20 & \bestval{1.74}  \\ 
& & RMSE & 5.59     & 4.76   & 4.25   &  5.18  &  4.63 &  4.55 & \bestval{3.97}   \\ 
& & MAPE & 6.8\%  & 5.4\% & 5.0\% & 5.5\% & 5.43\% & 5.2\% &  \bestval{3.9}\% \\ 
\cline{2-10}
& \multirow{3}{*}{1 hour}
  & MAE  & 2.88  & 3.38  &  2.93 & 3.28  &  2.46 & 2.37 & \bestval{2.07}   \\ 
& & RMSE & 5.59  & 6.50  & 5.44  & 7.08  &   4.98 &  4.96 & \bestval{4.74}   \\ 
& & MAPE & 6.8\%  & 8.3\% & 6.5\% & 8.0\%   &  5.89\%  &  5.7\% &   \bestval{4.9\%}   \\ 

\bottomrule

\end{tabular}
}
\end{table}

We compare \gcrnn{}\footnote{The source code is available at \url{https://github.com/liyaguang/DCRNN}.} with widely used time series regression models, including
(1) HA: Historical Average, which models the traffic flow as a seasonal process, and uses weighted average of previous seasons as the prediction;
(2) ARIMA$_{kal}$: Auto-Regressive Integrated Moving Average model with Kalman filter which is widely used in time series prediction; 
(3) VAR: Vector Auto-Regression~\citep{hamilton1994time}.
(4) SVR: Support Vector Regression which uses linear support vector machine for the regression task; 
The following deep neural network based approaches are also included:
(5) Feed forward Neural network (FNN): Feed forward neural network with two hidden layers and L2 regularization.
(6) Recurrent Neural Network with fully connected LSTM hidden units (FC-LSTM)~\citep{sutskever2014sequence}.

All neural network based approaches are implemented using Tensorflow~\citep{abadi2016tensorflow}, and trained using the Adam optimizer with learning rate annealing.
The best hyperparameters are chosen using the Tree-structured Parzen Estimator (TPE)~\citep{bergstra2011algorithms} on the validation dataset. 
Detailed parameter settings for \gcrnn{} as well as baselines are available in Appendix~\ref{sec:detailed_exp_settings}.

\vspace{-0.1in}
\subsection{Traffic Forecasting Performance Comparison}
Table~\ref{tab:la_comparison} shows the comparison of different approaches for 15 minutes, 30 minutes and 1 hour ahead forecasting on both datasets.
These methods are evaluated based on three commonly used metrics in traffic forecasting, including
(1)  Mean Absolute Error (MAE), (2) Mean Absolute Percentage Error (MAPE),  and (3) Root Mean Squared Error (RMSE).
Missing values are excluded in calculating these metrics. Detailed formulations of these metrics are provided in Appendix~\ref{sec:metrics}.
We observe the following phenomenon in both of these datasets.
(1) RNN-based methods, including FC-LSTM and \gcrnn{}, generally outperform other baselines which emphasizes the importance of modeling the temporal dependency.  
(2) \gcrnn{} achieves the best performance regarding all the metrics for all forecasting horizons, which suggests the effectiveness of spatiotemporal dependency modeling.
(3) Deep neural network based methods including FNN, FC-LSTM and \gcrnn{}, tend to have better performance than linear baselines for long-term forecasting, e.g., 1 hour ahead.  This is because the temporal dependency becomes increasingly non-linear with the growth of the horizon. 
Besides, as the historical average method does not depend on short-term data, its performance is invariant to the small increases in the forecasting horizon.

Note that, traffic forecasting on the METR-LA (Los Angeles, which is known for its complicated traffic conditions) dataset is more challenging than that in the PEMS-BAY (Bay Area) dataset.
Thus we use METR-LA as the default dataset for following experiments.

\vspace{-0.1in}
\subsection{Effect of spatial dependency modeling}
\vspace{-0.1in}
\begin{figure*}[tp]
    \centering
	\begin{minipage}{.48\textwidth}
	\centering
	\includegraphics[width=0.9\linewidth]{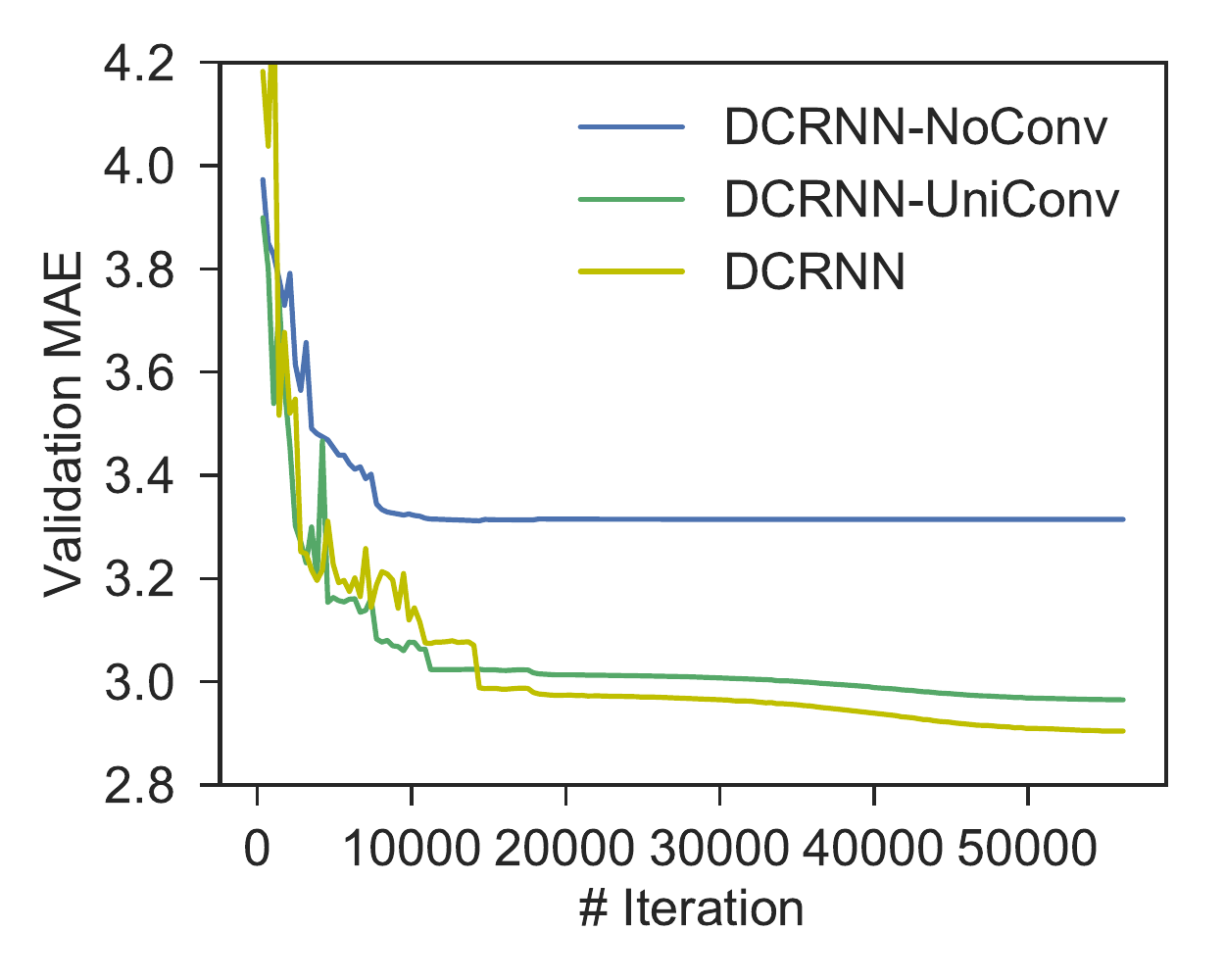}
	\caption{Learning curve for \gcrnn{} and \gcrnn{} without diffusion convolution. Removing diffusion convolution results in much higher validation error. Moreover, \gcrnn{} with bi-directional random walk achieves the lowest validation error.}
	\label{fig:spatial_dependency}
	\end{minipage}
	\hspace{0.01\textwidth}
	\begin{minipage}{.48\textwidth}
	\centering
    \includegraphics[width=0.9\linewidth]{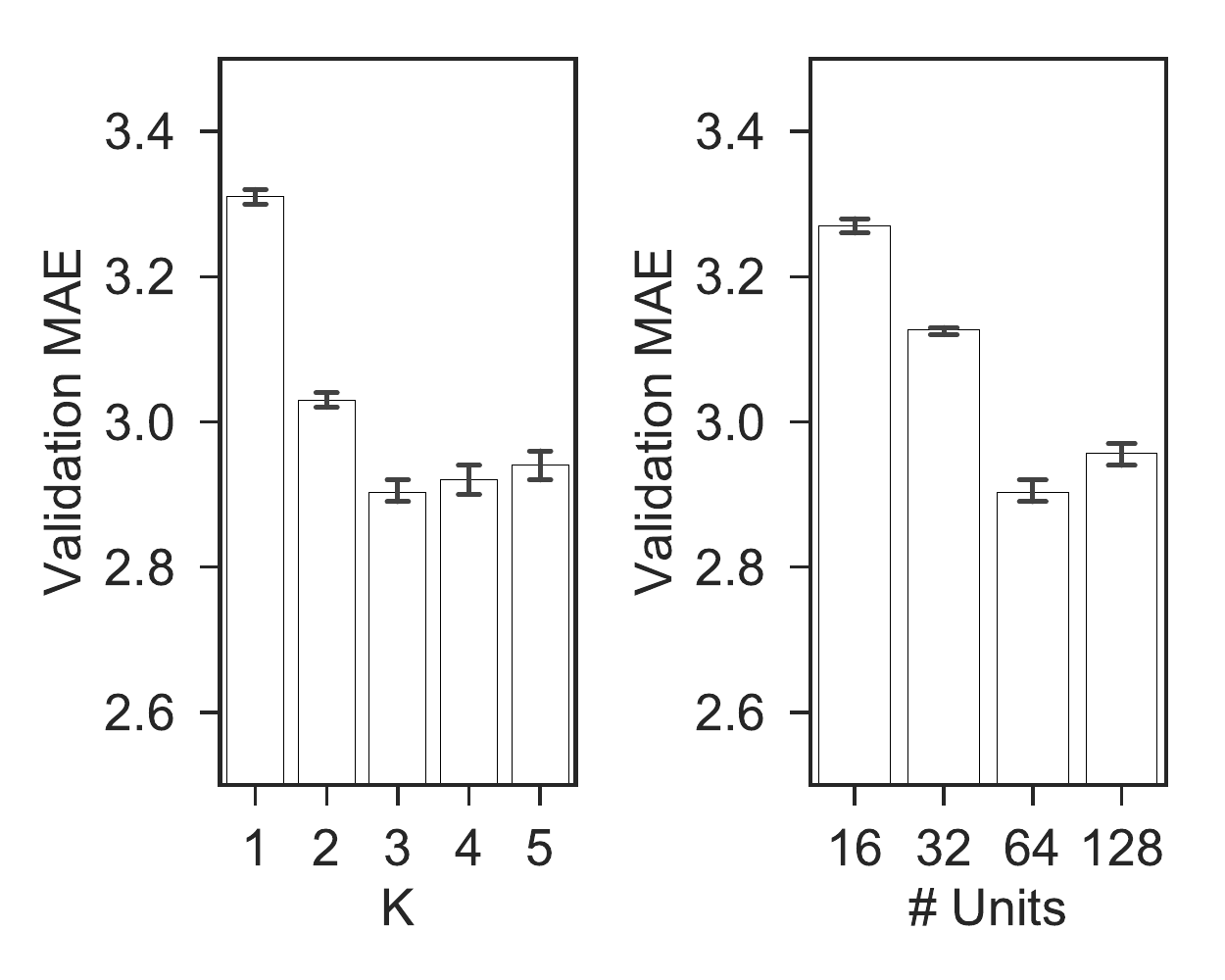}
	\caption{Effects of K and the number of units in each layer of \gcrnn{}. K corresponds to the reception field width of the filter, and the number of units corresponds to the number of filters.}
	\label{fig:k_effect}
	\end{minipage} 
\end{figure*}

\begin{table}[tp]
\centering
\caption{Performance comparison for \gcrnn{} and GCRNN on the METRA-LA dataset.}
\label{tab:dcrnn_vs_gcrnn}
\resizebox{0.85\columnwidth}{!}{
\begin{tabular}{c|ccc|ccc|ccc}
\toprule
      & \multicolumn{3}{c}{15 min} & \multicolumn{3}{c}{30 min} & \multicolumn{3}{c}{1 hour} \\ 
      \midrule
      & MAE     & RMSE   & MAPE    & MAE     & RMSE   & MAPE     & MAE    & RMSE   & MAPE     \\
      \hline
DCRNN & \bestval{2.77}    & \bestval{5.38}   & \bestval{7.3\%}   & \bestval{3.15}    & \bestval{6.45}   & \bestval{8.8\%}   & \bestval{3.60}   & \bestval{7.60}   & \bestval{10.5\%}   \\
GCRNN & 2.80    & 5.51   & 7.5\%   & 3.24    & 6.74   & 9.0\%   & 3.81   & 8.16   & 10.9\%  \\
\bottomrule
\end{tabular}
}
\end{table}

To further investigate the effect of spatial dependency modeling, we compare \gcrnn{} with the following variants:
(1) \gcrnn{}-NoConv, which ignores spatial dependency by replacing the transition matrices in the diffusion convolution (Equation~\ref{eq:gconv}) with identity matrices. This essentially means the forecasting of a sensor can be only be inferred from its own historical readings;
(2) \gcrnn{}-UniConv, which only uses the forward random walk transition matrix for diffusion convolution;
%
Figure~\ref{fig:spatial_dependency} shows the learning curves of these three models with roughly the same number of parameters.
Without diffusion convolution, \gcrnn{}-NoConv has much higher validation error. Moreover, \gcrnn{} achieves the lowest validation error which shows the effectiveness of using bidirectional random walk.
The intuition is that the bidirectional random walk gives the model the ability and flexibility to capture the influence from both the upstream and the downstream traffic.

To investigate the effect of graph construction, we construct a undirected graph by setting $\widehat{W}_{ij} = \widehat{W}_{ji} = \max(W_{ij}, W_{ji})$, where $\widehat{\M{W}}$ is the new symmetric weight matrix.
Then we develop a variant of \gcrnn{} denotes GCRNN, which uses the sequence to sequence learning with \emph{ChebNet graph convolution} (Equation~\ref{eq:chebnet}) with roughly the same amount of parameters.
Table~\ref{tab:dcrnn_vs_gcrnn} shows the comparison between \gcrnn{} and GCRNN in the METR-LA dataset.
\gcrnn{} consistently outperforms GCRNN. The intuition is that directed graph better captures the asymmetric correlation between traffic sensors.  
\begin{figure*}[tp]
    \centering
	\begin{minipage}{.43\textwidth}
	\centering
	\includegraphics[width=\linewidth]{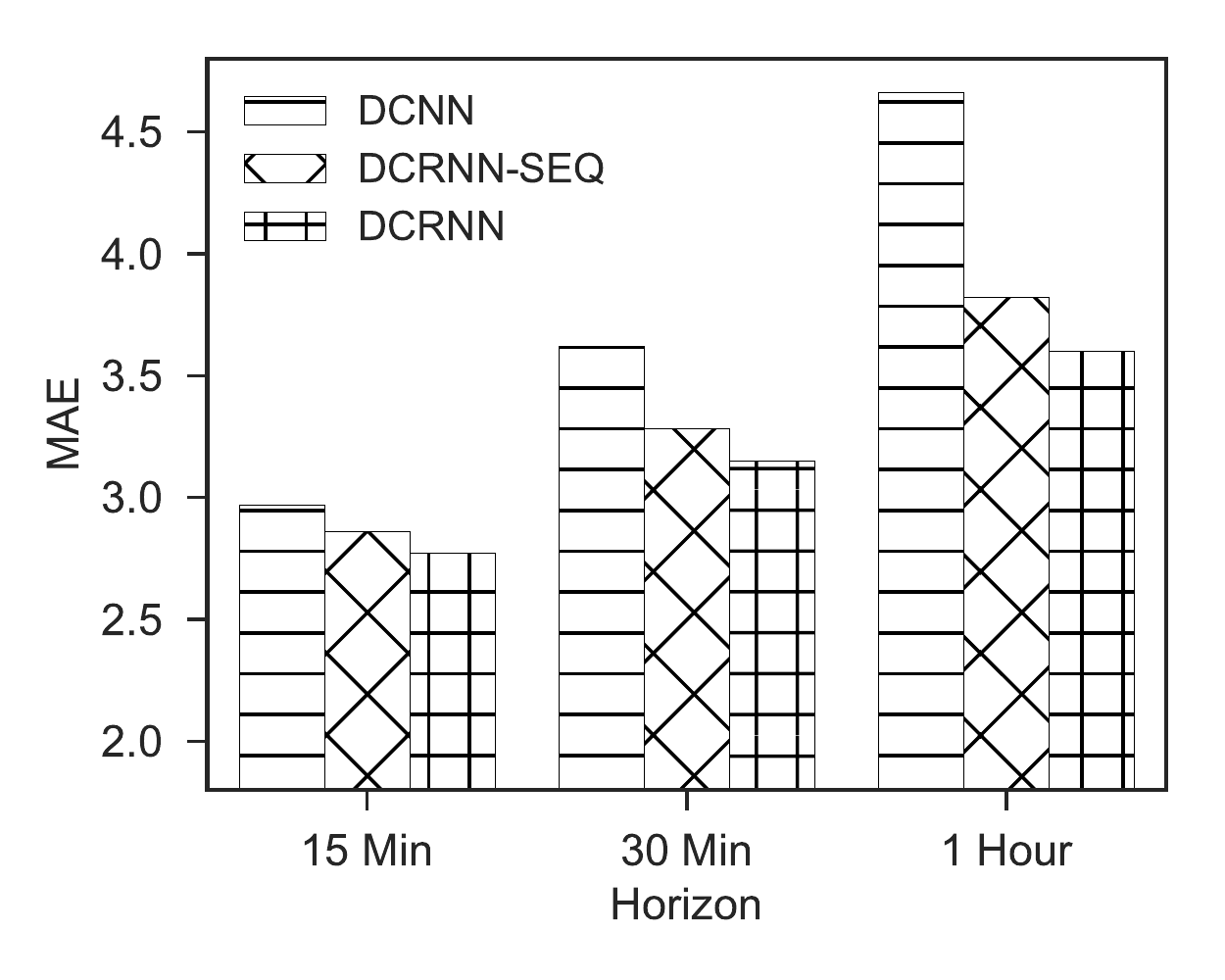}
    \caption{Performance comparison for different \gcrnn{} variants. \gcrnn{}, with the sequence to sequence framework and scheduled sampling, achieves the lowest MAE on the validation dataset. The advantage becomes more clear with the increase of the forecasting horizon.}
    \label{fig:seq2seq_comparison}
	\end{minipage}
	\hspace{0.01\textwidth}
	\begin{minipage}{.52\textwidth}
	\centering
	\includegraphics[width=\linewidth]{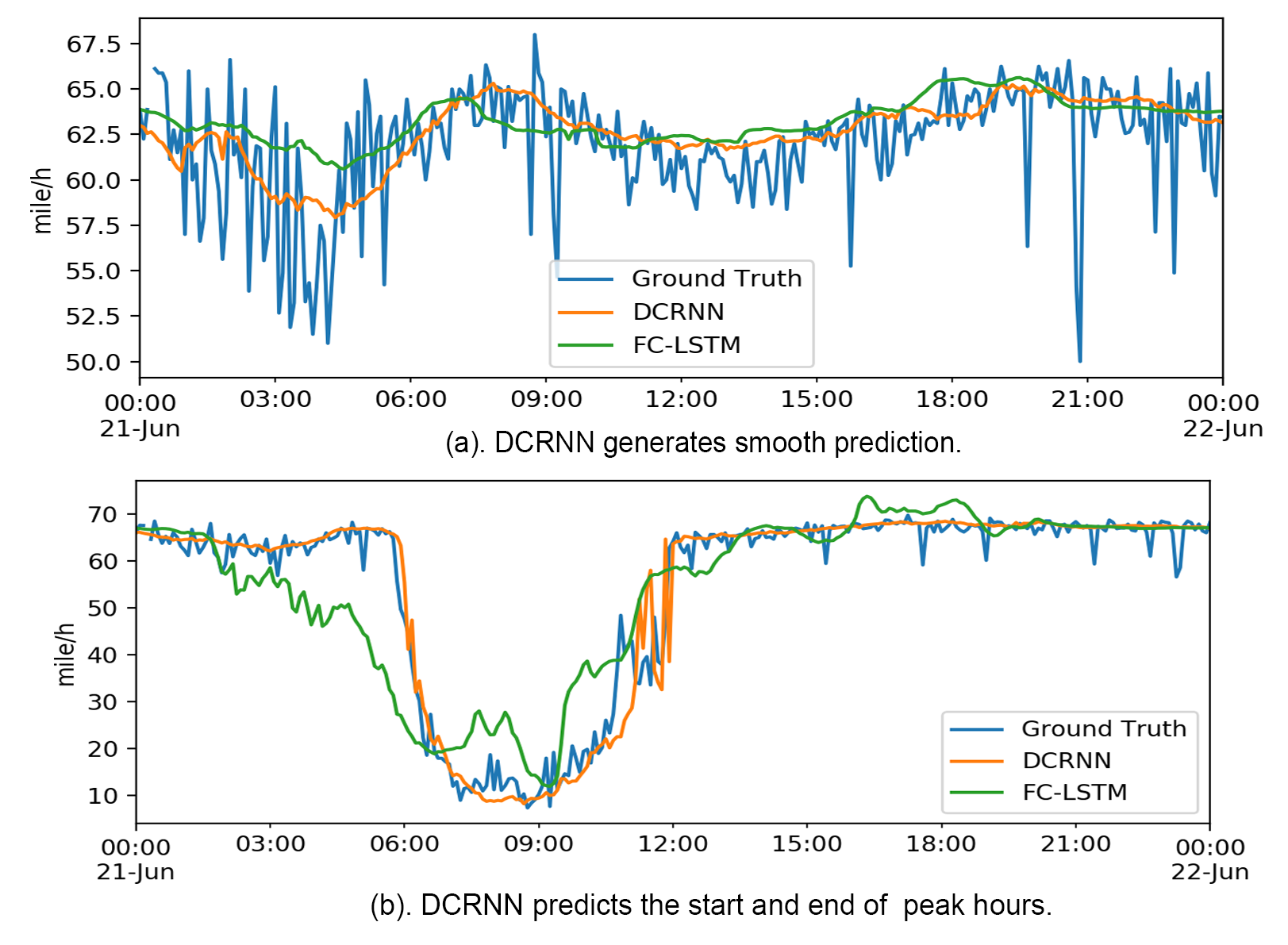}
	\caption{Traffic time series forecasting visualization. \gcrnn{} generates smooth prediction and is usually better at predict the start and end of peak hours.}
	\label{fig:traffic_visualization}
	\end{minipage} 
\end{figure*}
Figure~\ref{fig:k_effect} shows the effects of different parameters. 
$K$ roughly corresponds to the size of filters' reception fields while the number of units corresponds to the number of filters. 
Larger $K$ enables the model to capture broader spatial dependency at the cost of increasing learning complexity. We observe that with the increase of $K$, the error on the validation dataset first quickly decrease, and then slightly increase.
Similar behavior is observed for varying the number of units.
\vspace{-0.1in}
\subsection{Effect of temporal dependency modeling}

To evaluate the effect of temporal modeling including the sequence to sequence framework as well as the scheduled sampling mechanism, we further design three variants of \gcrnn{}: 
(1) DCNN: in which we concatenate the historical observations as a fixed length vector and feed it into stacked diffusion convolutional layers to predict the future time series.
We train a single model for one step ahead prediction, and feed the previous prediction into the model as input to perform multiple steps ahead prediction.
(2) \gcrnn{}-SEQ: which uses the encoder-decoder sequence to sequence learning framework to perform multiple steps ahead forecasting.
(3) \gcrnn{}: similar to \gcrnn{}-SEQ except for adding scheduled sampling.

Figure~\ref{fig:seq2seq_comparison} shows the comparison of those four methods with regards to MAE for different forecasting horizons. We observe that:
(1) \gcrnn{}-SEQ outperforms DCNN by a large margin which conforms the importance of modeling temporal dependency.
(2) \gcrnn{} achieves the best result, and its superiority becomes more evident with the increase of the forecasting horizon. 
This is mainly because the model is trained to deal with its mistakes during multiple steps ahead prediction and thus suffers less from the problem of error propagation. 
We also train a model that always been fed its output as input for multiple steps ahead prediction. However, its performance is much worse than all the three variants which emphasizes the importance of scheduled sampling.

\vspace{-0.1in}
\subsection{Model Interpretation}
\vspace{-0.1in}
\begin{figure*}[tp]
    \centering
	\includegraphics[width=\linewidth]{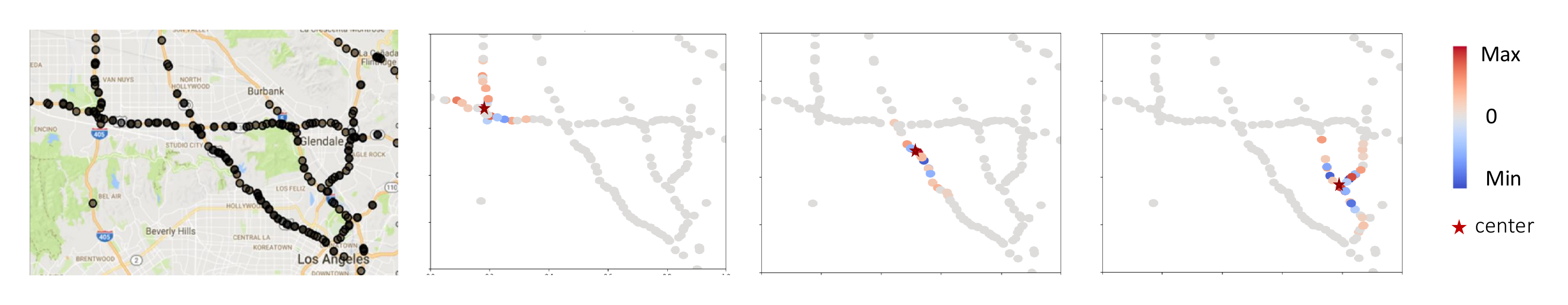}
	\caption{Visualization of learned localized filters centered at different nodes with $K=3$ on the METR-LA dataset. The star denotes the center, and the colors represent the weights. We observe that weights are localized around the center, and diffuse alongside the road network.}
	\label{fig:filter_visualization}
\end{figure*}

To better understand the model, we visualize forecasting results as well as learned filters. 
Figure~\ref{fig:traffic_visualization} shows the visualization of 1 hour ahead forecasting. We have the following observations:
(1) \gcrnn{} generates smooth prediction of the mean when small oscillation exists in the traffic speeds (Figure~\ref{fig:traffic_visualization}(a)). 
This reflects the robustness of the model.
(2) \gcrnn{} is more likely to accurately predict abrupt changes in the traffic speed than baseline methods (e.g., FC-LSTM).
As shown in Figure~\ref{fig:traffic_visualization}(b), \gcrnn{} predicts the start and the end of the peak hours. This is because \gcrnn{} captures the spatial dependency, and is able to utilize the speed changes in neighborhood sensors for more accurate forecasting. 
Figure~\ref{fig:filter_visualization} visualizes examples of learned filters centered at different nodes. The star denotes the center, and colors denote the weights. We can observe that (1) weights are well localized around the center, and (2) the weights diffuse based on road network distance. 
More visualizations are provided in Appendix~\ref{sec:model_visualization}.


\vspace{-0.1in}
\section{Conclusion}
\vspace{-0.1in}
In this paper, we formulated the traffic prediction on road network as a spatiotemporal forecasting problem, and proposed the \emph{diffusion convolutional recurrent neural network} that captures the spatiotemporal dependencies.
Specifically, we use bidirectional graph random walk to model spatial dependency and recurrent neural network to capture the temporal dynamics. 
We further integrated the encoder-decoder architecture and the scheduled sampling technique to improve the performance for long-term forecasting. 
When evaluated on two large-scale real-world traffic datasets, our approach obtained significantly better prediction than baselines. 
For future work, we will investigate the following two aspects (1) applying the proposed model to other spatial-temporal forecasting tasks; (2) modeling the spatiotemporal dependency when the underlying graph structure is evolving, e.g., the K nearest neighbor graph for moving objects.
\vspace{-0.1in}
\subsubsection*{Acknowledgments}
This research has been funded in part by NSF grants CNS-1461963, IIS-1254206, IIS-1539608, Caltrans-65A0533, the USC Integrated Media Systems Center (IMSC), and the USC METRANS Transportation Center. 
Any opinions, findings, and conclusions or recommendations expressed in this material are those of the authors and do not necessarily reflect the views of any of the sponsors such as NSF. 
Also, the authors would like to thank Shang-Hua Teng, Dehua Cheng and Siyang Li for helpful discussions and comments.

\bibliography{main}

\begin{thebibliography}{38}
\providecommand{\natexlab}[1]{#1}
\providecommand{\url}[1]{\texttt{#1}}
\expandafter\ifx\csname urlstyle\endcsname\relax
  \providecommand{\doi}[1]{doi: #1}\else
  \providecommand{\doi}{doi: \begingroup \urlstyle{rm}\Url}\fi

\bibitem[Abadi et~al.(2016)]{abadi2016tensorflow}
Mart{\'\i}n Abadi et~al.
\newblock Tensorflow: Large-scale machine learning on heterogeneous distributed
  systems.
\newblock \emph{arXiv preprint arXiv:1603.04467}, 2016.

\bibitem[Atwood \& Towsley(2016)Atwood and Towsley]{atwood2016diffusion}
James Atwood and Don Towsley.
\newblock Diffusion-convolutional neural networks.
\newblock In \emph{Advances in Neural Information Processing Systems}, pp.\
  1993--2001, 2016.

\bibitem[Bengio et~al.(2015)Bengio, Vinyals, Jaitly, and
  Shazeer]{bengio2015scheduled}
Samy Bengio, Oriol Vinyals, Navdeep Jaitly, and Noam Shazeer.
\newblock Scheduled sampling for sequence prediction with recurrent neural
  networks.
\newblock In \emph{NIPS}, pp.\  1171--1179, 2015.

\bibitem[Bergstra et~al.(2011)Bergstra, Bardenet, Bengio, and
  K{\'e}gl]{bergstra2011algorithms}
James~S Bergstra, R{\'e}mi Bardenet, Yoshua Bengio, and Bal{\'a}zs K{\'e}gl.
\newblock Algorithms for hyper-parameter optimization.
\newblock In \emph{Advances in Neural Information Processing Systems}, pp.\
  2546--2554, 2011.

\bibitem[Bruna et~al.(2014)Bruna, Zaremba, Szlam, and LeCun]{bruna2013spectral}
Joan Bruna, Wojciech Zaremba, Arthur Szlam, and Yann LeCun.
\newblock Spectral networks and locally connected networks on graphs.
\newblock In \emph{ICLR}, 2014.

\bibitem[Cai et~al.(2016)Cai, Wang, Lu, Chen, Ding, and
  Sun]{cai2016spatiotemporal}
Pinlong Cai, Yunpeng Wang, Guangquan Lu, Peng Chen, Chuan Ding, and Jianping
  Sun.
\newblock A spatiotemporal correlative k-nearest neighbor model for short-term
  traffic multistep forecasting.
\newblock \emph{Transportation Research Part C: Emerging Technologies},
  62:\penalty0 21--34, 2016.

\bibitem[Cascetta(2013)]{cascetta2013transportation}
Ennio Cascetta.
\newblock \emph{Transportation systems engineering: theory and methods},
  volume~49.
\newblock Springer Science \& Business Media, 2013.

\bibitem[Cheng et~al.(2015)Cheng, Cheng, Liu, Peng, and
  Teng]{cheng2015efficient}
Dehua Cheng, Yu~Cheng, Yan Liu, Richard Peng, and Shang-Hua Teng.
\newblock Efficient sampling for gaussian graphical models via spectral
  sparsification.
\newblock In \emph{Conference on Learning Theory}, pp.\  364--390, 2015.

\bibitem[Cheng et~al.(2017)Cheng, Zhang, Zhou, and Xu]{cheng2017deeptransport}
Xingyi Cheng, Ruiqing Zhang, Jie Zhou, and Wei Xu.
\newblock Deeptransport: Learning spatial-temporal dependency for traffic
  condition forecasting.
\newblock \emph{arXiv preprint arXiv:1709.09585}, 2017.

\bibitem[Chung et~al.(2014)Chung, Gulcehre, Cho, and
  Bengio]{chung2014empirical}
Junyoung Chung, Caglar Gulcehre, KyungHyun Cho, and Yoshua Bengio.
\newblock Empirical evaluation of gated recurrent neural networks on sequence
  modeling.
\newblock \emph{arXiv preprint arXiv:1412.3555}, 2014.

\bibitem[Defferrard et~al.(2016)Defferrard, Bresson, and
  Vandergheynst]{defferrard2016convolutional}
Micha{\"e}l Defferrard, Xavier Bresson, and Pierre Vandergheynst.
\newblock Convolutional neural networks on graphs with fast localized spectral
  filtering.
\newblock In \emph{NIPS}, pp.\  3837--3845, 2016.

\bibitem[Deng et~al.(2016)Deng, Shahabi, Demiryurek, Zhu, Yu, and
  Liu]{DengSDZYL16}
Dingxiong Deng, Cyrus Shahabi, Ugur Demiryurek, Linhong Zhu, Rose Yu, and Yan
  Liu.
\newblock Latent space model for road networks to predict time-varying traffic.
\newblock In \emph{{SIGKDD}}, pp.\  1525--1534, 2016.

\bibitem[Drew(1968)]{drew1968traffic}
Donald~R Drew.
\newblock Traffic flow theory and control.
\newblock Technical report, 1968.

\bibitem[Fusco et~al.(2016)Fusco, Colombaroni, and Isaenko]{fusco2016short}
Gaetano Fusco, Chiara Colombaroni, and Natalia Isaenko.
\newblock Short-term speed predictions exploiting big data on large urban road
  networks.
\newblock \emph{Transportation Research Part C: Emerging Technologies},
  73:\penalty0 183--201, 2016.

\bibitem[Gehring et~al.(2017)Gehring, Auli, Grangier, Yarats, and
  Dauphin]{gehring2017convolutional}
Jonas Gehring, Michael Auli, David Grangier, Denis Yarats, and Yann~N Dauphin.
\newblock Convolutional sequence to sequence learning.
\newblock In \emph{ICML}, 2017.

\bibitem[Grover \& Leskovec(2016)Grover and Leskovec]{grover2016node2vec}
Aditya Grover and Jure Leskovec.
\newblock node2vec: Scalable feature learning for networks.
\newblock In \emph{Proceedings of the 22nd ACM SIGKDD international conference
  on Knowledge discovery and data mining}, pp.\  855--864. ACM, 2016.

\bibitem[Hamilton(1994)]{hamilton1994time}
James~Douglas Hamilton.
\newblock \emph{Time series analysis}, volume~2.
\newblock Princeton university press Princeton, 1994.

\bibitem[Hechtlinger et~al.(2017)Hechtlinger, Chakravarti, and
  Qin]{hechtlinger2017generalization}
Yotam Hechtlinger, Purvasha Chakravarti, and Jining Qin.
\newblock A generalization of convolutional neural networks to graph-structured
  data.
\newblock \emph{arXiv preprint arXiv:1704.08165}, 2017.

\bibitem[Jagadish et~al.(2014)Jagadish, Gehrke, Labrinidis, Papakonstantinou,
  Patel, Ramakrishnan, and Shahabi]{Jagadish2014}
H.~V. Jagadish, Johannes Gehrke, Alexandros Labrinidis, Yannis
  Papakonstantinou, Jignesh~M. Patel, Raghu Ramakrishnan, and Cyrus Shahabi.
\newblock Big data and its technical challenges.
\newblock \emph{Commun. ACM}, 57\penalty0 (7):\penalty0 86--94, July 2014.

\bibitem[Kipf \& Welling(2017)Kipf and Welling]{kipf2016semi}
Thomas~N Kipf and Max Welling.
\newblock Semi-supervised classification with graph convolutional networks.
\newblock In \emph{International Conference on Learning Representations
  (ICLR)}, 2017.

\bibitem[Laptev et~al.(2017)Laptev, Yosinski, Li, and Smyl]{laptev2017time}
Nikolay Laptev, Jason Yosinski, Li~Erran Li, and Slawek Smyl.
\newblock Time-series extreme event forecasting with neural networks at {Uber}.
\newblock In \emph{Int. Conf. on Machine Learning Time Series Workshop}, 2017.

\bibitem[Lippi et~al.(2013)Lippi, Bertini, and Frasconi]{lippi2013short}
Marco Lippi, Marco Bertini, and Paolo Frasconi.
\newblock Short-term traffic flow forecasting: An experimental comparison of
  time-series analysis and supervised learning.
\newblock \emph{ITS, IEEE Transactions on}, 14\penalty0 (2):\penalty0 871--882,
  2013.

\bibitem[Liu et~al.(2011)Liu, Zheng, Chawla, Yuan, and
  Xing]{liu2011discovering}
Wei Liu, Yu~Zheng, Sanjay Chawla, Jing Yuan, and Xie Xing.
\newblock Discovering spatio-temporal causal interactions in traffic data
  streams.
\newblock In \emph{SIGKDD}, pp.\  1010--1018. ACM, 2011.

\bibitem[Lv et~al.(2015)Lv, Duan, Kang, Li, and Wang]{lv2015traffic}
Yisheng Lv, Yanjie Duan, Wenwen Kang, Zhengxi Li, and Fei-Yue Wang.
\newblock Traffic flow prediction with big data: A deep learning approach.
\newblock \emph{ITS, IEEE Transactions on}, 16\penalty0 (2):\penalty0 865--873,
  2015.

\bibitem[Ma et~al.(2017)Ma, Dai, He, Ma, Wang, and Wang]{ma2017learning}
Xiaolei Ma, Zhuang Dai, Zhengbing He, Jihui Ma, Yong Wang, and Yunpeng Wang.
\newblock Learning traffic as images: a deep convolutional neural network for
  large-scale transportation network speed prediction.
\newblock \emph{Sensors}, 17\penalty0 (4):\penalty0 818, 2017.

\bibitem[Perozzi et~al.(2014)Perozzi, Al-Rfou, and Skiena]{perozzi2014deepwalk}
Bryan Perozzi, Rami Al-Rfou, and Steven Skiena.
\newblock Deepwalk: Online learning of social representations.
\newblock In \emph{Proceedings of the 20th ACM SIGKDD international conference
  on Knowledge discovery and data mining}, pp.\  701--710. ACM, 2014.

\bibitem[Seo et~al.(2016)Seo, Defferrard, Vandergheynst, and
  Bresson]{seo2016structured}
Youngjoo Seo, Micha{\"e}l Defferrard, Pierre Vandergheynst, and Xavier Bresson.
\newblock Structured sequence modeling with graph convolutional recurrent
  networks.
\newblock \emph{arXiv preprint arXiv:1612.07659}, 2016.

\bibitem[Shuman et~al.(2013)Shuman, Narang, Frossard, Ortega, and
  Vandergheynst]{shuman2013emerging}
David~I Shuman, Sunil~K Narang, Pascal Frossard, Antonio Ortega, and Pierre
  Vandergheynst.
\newblock The emerging field of signal processing on graphs: Extending
  high-dimensional data analysis to networks and other irregular domains.
\newblock \emph{IEEE Signal Processing Magazine}, 30\penalty0 (3):\penalty0
  83--98, 2013.

\bibitem[Sutskever et~al.(2014)Sutskever, Vinyals, and
  Le]{sutskever2014sequence}
Ilya Sutskever, Oriol Vinyals, and Quoc~V Le.
\newblock Sequence to sequence learning with neural networks.
\newblock In \emph{NIPS}, pp.\  3104--3112, 2014.

\bibitem[Teng et~al.(2016)]{teng2016scalable}
Shang-Hua Teng et~al.
\newblock Scalable algorithms for data and network analysis.
\newblock \emph{Foundations and Trends{\textregistered} in Theoretical Computer
  Science}, 12\penalty0 (1--2):\penalty0 1--274, 2016.

\bibitem[Vlahogianni et~al.(2014)Vlahogianni, Karlaftis, and
  Golias]{vlahogianni2014short}
Eleni~I Vlahogianni, Matthew~G Karlaftis, and John~C Golias.
\newblock Short-term traffic forecasting: Where we are and where we’re going.
\newblock \emph{Transportation Research Part C: Emerging Technologies},
  43:\penalty0 3--19, 2014.

\bibitem[Wu \& Tan(2016)Wu and Tan]{wu2016short}
Yuankai Wu and Huachun Tan.
\newblock Short-term traffic flow forecasting with spatial-temporal correlation
  in a hybrid deep learning framework.
\newblock \emph{arXiv preprint arXiv:1612.01022}, 2016.

\bibitem[Xie et~al.(2010)Xie, Zhao, Sun, and Chen]{xie2010gaussian}
Yuanchang Xie, Kaiguang Zhao, Ying Sun, and Dawei Chen.
\newblock Gaussian processes for short-term traffic volume forecasting.
\newblock \emph{Transportation Research Record: Journal of the Transportation
  Research Board}, \penalty0 (2165):\penalty0 69--78, 2010.

\bibitem[Yu et~al.(2017{\natexlab{a}})Yu, Yin, and Zhu]{yu2017spatio}
Bing Yu, Haoteng Yin, and Zhanxing Zhu.
\newblock Spatio-temporal graph convolutional neural network: A deep learning
  framework for traffic forecasting.
\newblock \emph{arXiv preprint arXiv:1709.04875}, 2017{\natexlab{a}}.

\bibitem[Yu et~al.(2016)Yu, Rao, and Dhillon]{yu2016temporal}
Hsiang-Fu Yu, Nikhil Rao, and Inderjit~S Dhillon.
\newblock Temporal regularized matrix factorization for high-dimensional time
  series prediction.
\newblock In \emph{Advances in Neural Information Processing Systems}, pp.\
  847--855, 2016.

\bibitem[Yu et~al.(2017{\natexlab{b}})Yu, Li, Shahabi, Demiryurek, and
  Liu]{deeptraffic2017}
Rose Yu, Yaguang Li, Cyrus Shahabi, Ugur Demiryurek, and Yan Liu.
\newblock Deep learning: A generic approach for extreme condition traffic
  forecasting.
\newblock In \emph{SIAM International Conference on Data Mining (SDM)},
  2017{\natexlab{b}}.

\bibitem[Zhang et~al.(2016)Zhang, Zheng, Qi, Li, and Yi]{zhang2016dnn}
Junbo Zhang, Yu~Zheng, Dekang Qi, Ruiyuan Li, and Xiuwen Yi.
\newblock Dnn-based prediction model for spatio-temporal data.
\newblock In \emph{Proceedings of the 24th ACM SIGSPATIAL International
  Conference on Advances in Geographic Information Systems}, pp.\ ~92. ACM,
  2016.

\bibitem[Zhang et~al.(2017)Zhang, Zheng, and Qi]{zhang2017deep}
Junbo Zhang, Yu~Zheng, and Dekang Qi.
\newblock Deep spatio-temporal residual networks for citywide crowd flows
  prediction.
\newblock In \emph{AAAI}, pp.\  1655--1661, 2017.

\end{thebibliography}
\bibliographystyle{iclr2018_conference}

\clearpage

\section*{Appendix}
\renewcommand{\thesubsection}{\Alph{subsection}}
\subsection{Notation}
\begin{table}[htp]
\centering
\caption{Notation}
\label{tab:notation}
\begin{tabular}{l|l}
Name &   \\
\hline \hline
 $\graph{}$  & a graph \\
 $\vertices{}, v_i $ & nodes of a graph, $|\vertices{}|=N$  and the $i$-th node. \\
 $\edges{}$ & edges of a graph  \\
 $\M{W}, W_{ij}, $ & weight matrix of a graph and its entries \\
  $\M{D}, \M{D}_I, \M{D}_O$ & undirected degree matrix, In-degree/out-degree matrix  \\
 $\M{L}$ & normalized graph Laplacian  \\
  $\M{\Phi}, \M{\Lambda}$ & eigen-vector matrix and eigen-value matrix of $\M{L}$\\
 $\M{X},\hat{\M{X}} \in \real^{N \times P}$ & a graph signal, and the predicted graph signal. \\
 $\M{X}^{(t)} \in \real^{N \times P}$ & a graph signal at time $t$. \\
 $\M{H} \in \real^{N \times Q}$ & output of the diffusion convolutional layer. \\
$f_{\V{\theta}}, \V{\theta}$ & convolutional filter and its parameters.\\
$f_{\M{\Theta}}, \M{\Theta}$ & convolutional layer and its parameters.\\
 \hline
\end{tabular}
\end{table}
Table~\ref{tab:notation} summarizes the main notations used in  the paper.

\subsection{Efficient Calculation of Equation~\ref{eq:gconv}}
\label{sec:recursive_calculation}
Equation~\ref{eq:gconv} can be decomposed into two parts with the same time complexity, i.e., one part with $\M{D}_O^{-1} \M{W}$ and the other part with $\M{D}_I^{-1}\M{W}^\intercal$. Thus we will only show the time complexity of the first part.

Let $T_k(\V{x}) = \left(\M{D}_O^{-1} \M{W}\right)^k \V{x}$,
The first part of Equation~\ref{eq:gconv} can be rewritten as 
\begin{equation}
\label{eq:gconv2}
\sum_{k=0}^{K-1} \theta_k T_k(X_{:,p})
\end{equation}
As $T_{k+1}(\V{x}) = \M{D}_O^{-1} \M{W}\, T_k(\V{x})$ and $\M{D}_O^{-1} \M{W}$ is sparse, it is easy to see that Equation~\ref{eq:gconv2} can be calculated using $O(K)$ recursive sparse-dense matrix multiplication each with time complexity $O(|\edges|)$. Consequently, the time complexities of both Equation~\ref{eq:gconv} and Equation~\ref{eq:gconv2} are $O(K|\edges|)$.  For dense graph, we may use spectral sparsification~\citep{cheng2015efficient} to make it sparse.

\subsection{Relation with Spectral Graph Convolution}
\label{sec:gconv_relationship}
\proof
The spectral graph convolution utilizes the concept of normalized graph Laplacian
$\M{L}=\M{D}^{-\frac{1}{2}} (\M{D} - \M{W})\M{D}^{-\frac{1}{2}} =\mathbf{\Phi} \Lambda \mathbf{\Phi}^\intercal$.
ChebNet parametrizes $f_\theta$ to be a $K$ order polynomial of $\M{\Lambda}$, and calculates it using stable Chebyshev polynomial basis.
\begin{equation}
\M{X}_{:,p} \gconv{} f_\V{\theta} 
= \M{\Phi} \left(\sum_{k=0}^{K-1} \theta_{k}\M{\Lambda}^k \right)\mathbf{\Phi}^\intercal \M{X}_{:,p}
= \sum_{k=0}^{K-1} \theta_{k}\M{L}^k\M{X}_{:,p} 
= \sum_{k=0}^{K-1} \tilde{\theta}_{k}T_k(\tilde{\M{L}}) \M{X}_{:,p}
\label{eq:chebnet}
\end{equation}
where $T_0(x)=1, T_1(x)=x, T_k(x)=xT_{k-1}(x)-T_{k-2}(x)$ are the basis of the Cheyshev polynomial. 
Let $\lambda_{max}$ denote the largest eigenvalue of 
$\M{L}$, and $\tilde{\M{L}}=\frac{2}{\lambda_{max}} \M{L} - \M{I}$ represents a rescaling of the graph Laplacian that maps the eigenvalues from $[0, \lambda_{max}]$ to $[-1, 1]$ since  Chebyshev polynomial forms an orthogonal basis in $[-1, 1]$. 
Equation~\ref{eq:chebnet} can be considered as a polynomial of $\tilde{\M{L}}$ and we will show that the output of ChebNet Convolution is \emph{similar} to the output of diffusion convolution up to constant scaling factor. Assume $\lambda_{max} = 2$ and $\M{D}_I = \M{D}_O = \M{D}$ for undirected graph.
\begin{equation}
\tilde{\M{L}} = \M{D}^{-\frac{1}{2}} (\M{D} - \M{W})\M{D}^{-\frac{1}{2}} - \M{I} =  -\M{D}^{-\frac{1}{2}} \M{W} \M{D}^{-\frac{1}{2}} \sim -\M{D}^{-1} \M{W}
\label{eq:gconv_relationship}
\end{equation}
$\tilde{\M{L}}$ is \emph{similar} to the negative random walk transition matrix, thus the output of Equation~\ref{eq:chebnet} is also similar to the output of Equation~\ref{eq:gconv} up to constant scaling factor. \qed

\begin{figure*}[htp]
    \centering
	\includegraphics[width=\linewidth]{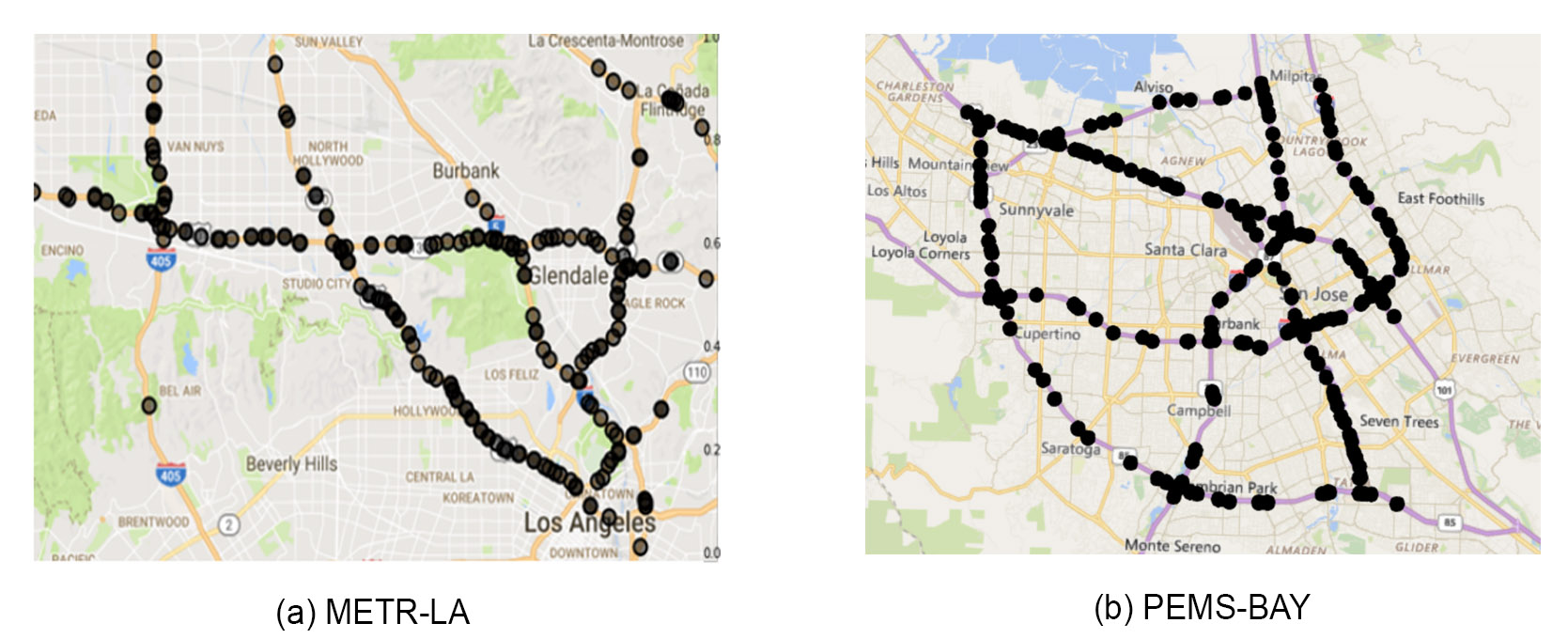}
	\caption{Sensor distribution of the METR-LA and PEMS-BAY dataset.}
	\label{fig:sensor_distribution}
\end{figure*}

\subsection{More Related Work and Discussion}
\label{sec:more_related_work}
\citet{xie2010gaussian} introduce a Gaussian processes (GPs) based method.
GPs are hard to scale to the large dataset and are generally not suitable for relatively long-term traffic prediction like 1 hour (i.e.,12 steps ahead), as the variance can be accumulated and becomes extremely large.

\citet{cai2016spatiotemporal} propose to use spatiotemporal nearest neighbor for traffic forecasting (ST-KNN). 
Though ST-KNN considers both the spatial and the temporal dependencies, it has the following drawbacks.
As shown in~\citet{fusco2016short}, ST-KNN performs independent forecasting for each individual road. The prediction of a road is a weighted combination of its own historical traffic speeds. This makes it hard for ST-KNN to fully utilize information from neighbors.
Besides, ST-KNN is a non-parametric approach and each road is modeled and calculated separately~\citep{cai2016spatiotemporal}, which makes it hard to generalize to unseen situations and to scale to large datasets. 
Finally, in ST-KNN, all the similarities are calculated using hand-designed metrics with few learnable parameters, and this may limit its representational power. 

\cite{cheng2017deeptransport} propose DeepTransport which models the spatial dependency by explicitly collecting certain number of upstream and downstream roads for each individual road and then conduct convolution on these roads respectively. 
Comparing with~\citet{cheng2017deeptransport}, \gcrnn{} models the spatial dependency in a more systematic way, i.e., generalizing convolution to the traffic sensor graph based on the diffusion nature of traffic. 
Besides, we derive \gcrnn{} from the property of random walk and show that the popular spectral convolution ChebNet is a special case of our method.

The proposed approach is also related to graph embedding techniques, e.g.,
Deepwalk~\citep{perozzi2014deepwalk}, node2vec~\citep{grover2016node2vec} which learn a low dimension representation for each node in the graph.
\gcrnn{} also learns a representation for each node. The learned representations capture both the spatial and the temporal dependency and at the same time are optimized with regarding to the objective, e.g., future traffic speeds.

\subsection{Detailed Experimental Settings}
\label{sec:detailed_exp_settings}
\paragraph{HA} Historical Average, which models the traffic flow as a seasonal process, and uses weighted average of previous seasons as the prediction. The period used is 1 week, and the prediction is based on aggregated data from previous weeks.
For example, the prediction for this Wednesday is the averaged traffic speeds from last four Wednesdays. As the historical average method does not depend on short-term data, its performance is invariant to the small increases in the forecasting horizon

\paragraph{ARIMA$_{kal}$} : Auto-Regressive Integrated Moving Average model with Kalman filter. The orders are (3, 0, 1), and the model is implemented using the \emph{statsmodel} python package.

\paragraph{VAR} Vector Auto-regressive model~\citep{hamilton1994time}. The number of lags is set to 3, and the model is implemented using the \emph{statsmodel} python package.

\paragraph{SVR} Linear Support Vector Regression, the penalty term $C=0.1$, the number of historical observation is 5.

The following deep neural network based approaches are also included.
\paragraph{FNN} Feed forward neural network with two hidden layers, each layer contains 256 units. The initial learning rate is $1e^{-3}$, and reduces to $\frac{1}{10}$ every 20 epochs starting at the 50th epochs. 
In addition, for all hidden layers, dropout with ratio 0.5 and L2 weight decay $1e^{-2}$ is used. The model is trained with batch size 64 and MAE as the loss function.  
Early stop is performed by monitoring the validation error.

\paragraph{FC-LSTM} The Encoder-decoder framework using LSTM with peephole~\citep{sutskever2014sequence}. Both the encoder and the decoder contain two  recurrent layers. 
In each recurrent layer, there are 256 LSTM units, L1 weight decay is $2e^{-5}$, L2 weight decay $5e^{-4}$. 
The model is trained with batch size 64 and loss function MAE. The initial learning rate is 1e-4 and reduces to $\frac{1}{10}$ every 10 epochs starting from the 20th epochs. Early stop is performed by monitoring the validation error.

\paragraph{\gcrnn{}}: Diffusion Convolutional Recurrent Neural Network. Both encoder and decoder contain two  recurrent layers. In each recurrent layer, there are 64 units, the initial learning rate is $1e^{-2}$, and reduces to $\frac{1}{10}$ every 10 epochs starting at the 20th epoch and early stopping on the validation dataset is used. 
Besides, the maximum steps of random walks, i.e., $K$, is set to 3.
For scheduled sampling, the thresholded inverse sigmoid function is used as the probability decay:
\begin{equation*}
\epsilon_i = \frac{\tau}{\tau+\exp{(i/\tau)}}
\end{equation*}
where $i$ is the number of iterations while $\tau$ are parameters to control the speed of convergence. $\tau$ is set to 3,000 in the experiments.
The implementation is available in \url{https://github.com/liyaguang/DCRNN}.

\subsubsection{Dataset}
We conduct experiments on two real-world large-scale datasets:
\begin{itemize}
	\item \textbf{METR-LA} This traffic dataset contains traffic information collected from loop detectors in the highway of Los Angeles County~\citep{Jagadish2014}.  We select 207 sensors and collect 4 months of data ranging from Mar 1st 2012 to Jun 30th 2012 for the experiment. 
	The total number of observed traffic data points is 6,519,002.
	\item \textbf{PEMS-BAY} This traffic dataset is collected by California Transportation Agencies (CalTrans) Performance Measurement System (PeMS). 
	We select 325 sensors in the Bay Area and collect 6 months of data ranging from Jan 1st 2017 to May 31th 2017 for the experiment.
	The total number of observed traffic data points is 16,937,179.
\end{itemize}
The sensor distributions of both datasets are visualized in~Figure~\ref{fig:sensor_distribution}.

In both of those datasets, we aggregate traffic speed readings into 5 minutes windows, and apply Z-Score normalization. 70\% of data is used for training, 20\% are used for testing while the remaining 10\% for validation. 
To construct the sensor graph, we compute the pairwise road network distances between sensors and build the adjacency matrix using thresholded Gaussian kernel~\citep{shuman2013emerging}. 
\begin{equation*}
W_{ij} = \exp{\left(-\frac{\mathrm{dist}({v_i, v_j})^2}{\sigma^2}\right)} \quad \text{if } \mathrm{dist}(v_i, v_j) \leq \kappa \text{, otherwise } 0
\end{equation*}
where $W_{ij}$ represents the edge weight between sensor $v_i$ and sensor $v_j$, $\mathrm{dist}({v_i, v_j})$ denotes the road network distance from sensor $v_i$ to sensor $v_j$. $\sigma$ is the standard deviation of distances and $\kappa$ is the threshold.

\subsubsection{Metrics}
\label{sec:metrics}
Suppose $\V{x}=x_1, \cdots, x_n$ represents the ground truth, $\hat{\V{x}}=\hat{x}_1, \cdots, \hat{x}_n$ represents the predicted values, and $\Omega$ denotes the indices of observed samples, the metrics are defined as follows.

Root Mean Square Error (RMSE)
\begin{equation*}
\text{RMSE}(\V{x}, \hat{\V{x}}) = \sqrt{\frac{1}{|\V{\Omega}|} \sum_{i \in \V{\Omega}} (x_i - \hat{x}_i)^2}
\end{equation*}

Mean Absolute Percentage Error (MAPE)
\begin{equation*}
\text{MAPE}(\V{x}, \hat{\V{x}}) = \frac{1}{|\V{\Omega}|} \sum_{i \in \V{\Omega}} \left|\frac{x_i - \hat{x}_i}{x_i}\right|
\end{equation*}

Mean Absolute Error (MAE)
\begin{equation*}
\text{MAE}(\V{x}, \hat{\V{x}}) = \frac{1}{|\V{\Omega}|} \sum_{i \in \V{\Omega}} \left|x_i - \hat{x}_i\right|
\end{equation*}

\subsection{Model Visualization}
\label{sec:model_visualization}
\begin{figure}[htp]
  \begin{center}
    \includegraphics[width=0.6\linewidth]{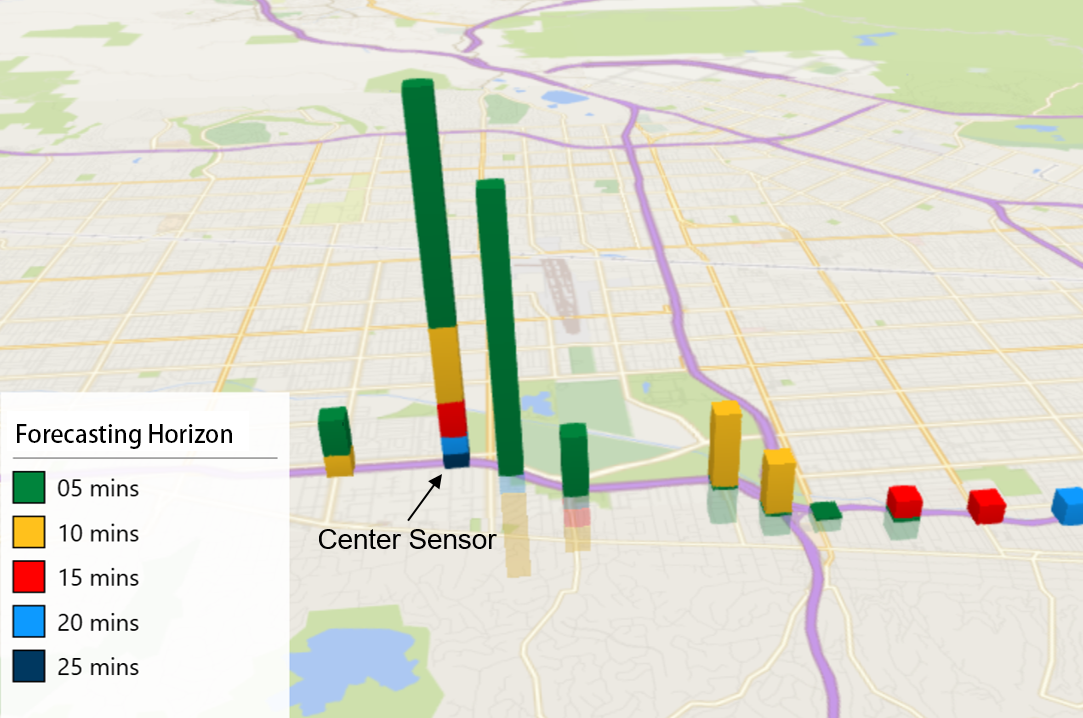}
  \end{center}
  \caption{Sensor correlations between the center sensor and its neighborhoods for different forecasting horizons. The correlations are estimated using regularized VAR. We observe that the correlations are localized and closer neighborhoods usually have
larger relevance, and the magnitude of correlation quickly
decay with the increase of distance which is consistent with the diffusion process on the graph. }
       \label{fig:sensor_correlation}
\end{figure}

\begin{figure*}[tp]
	\centering
	\includegraphics[width=0.8\linewidth]{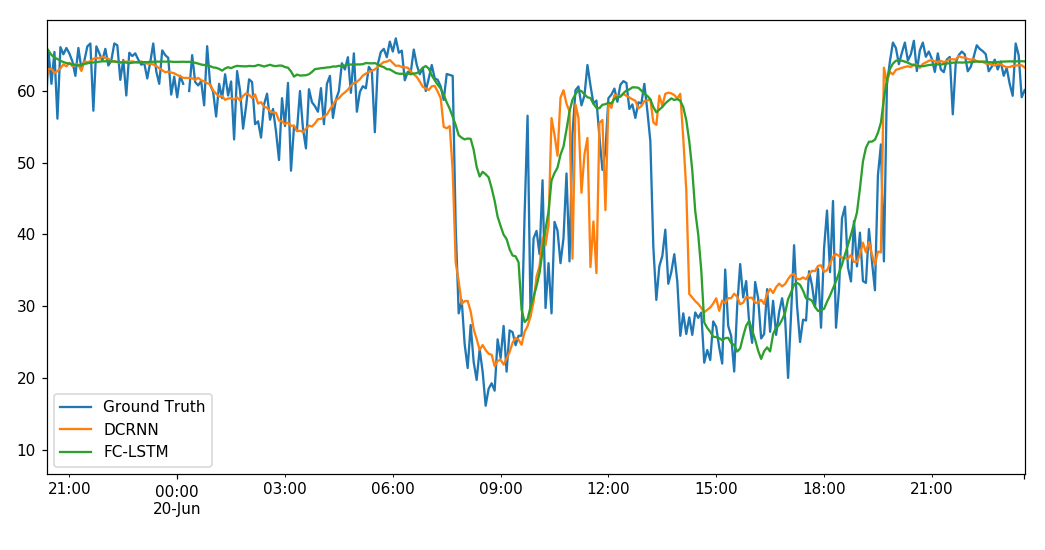}
    \includegraphics[width=0.8\linewidth]{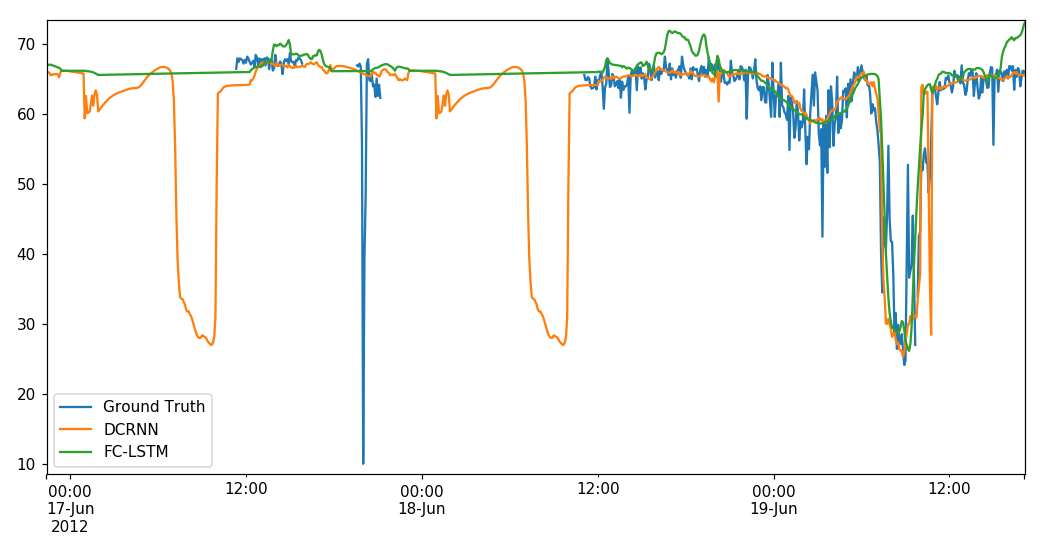}
    \includegraphics[width=0.8\linewidth]{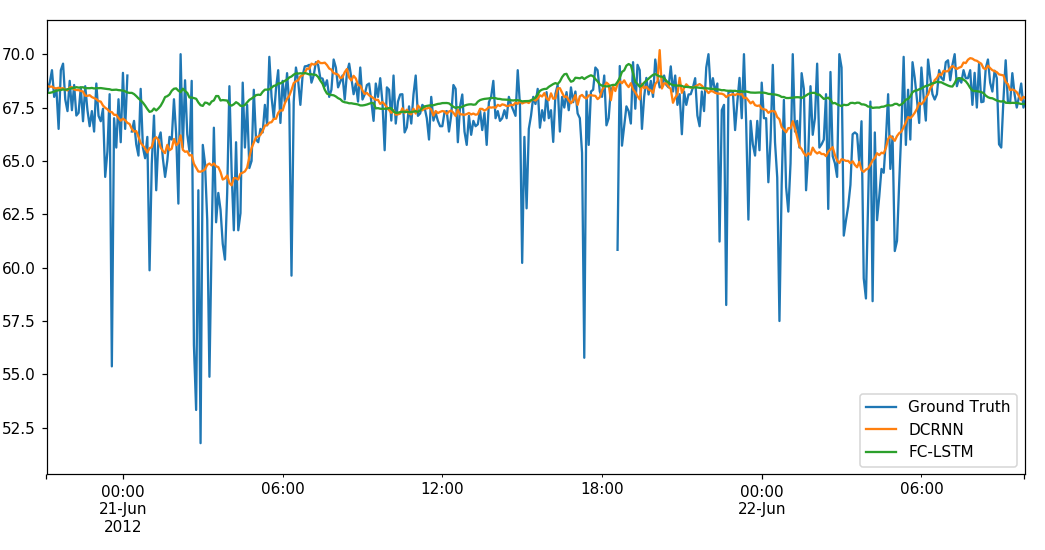}
    \includegraphics[width=0.8\linewidth]{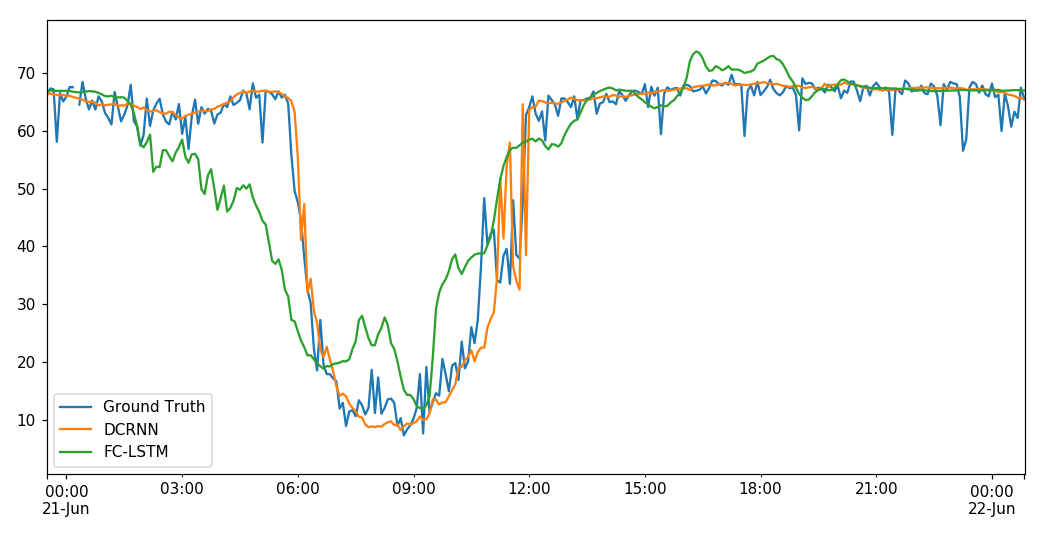}
	\caption{Traffic time series forecasting visualization.}
	\label{fig:traffic_visualization_large2}
\end{figure*}
\begin{figure*}[tp]
	\centering
	\includegraphics[width=0.8\linewidth]{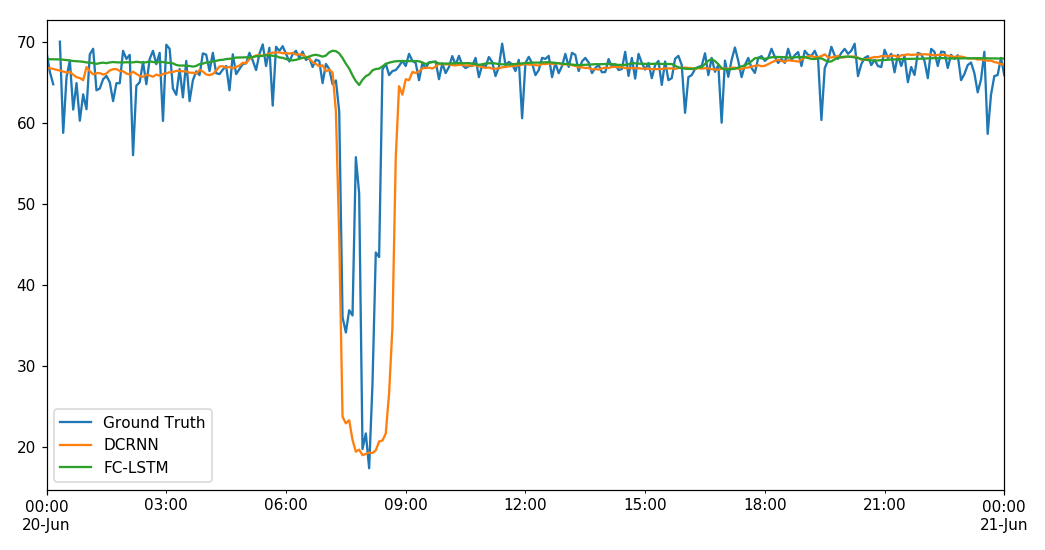}
    \includegraphics[width=0.8\linewidth]{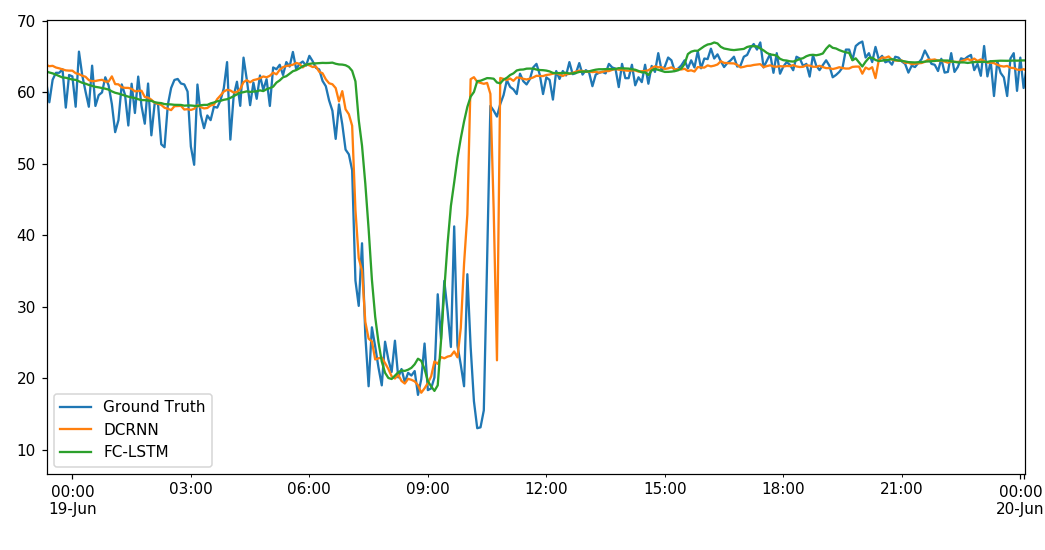}
    \includegraphics[width=0.8\linewidth]{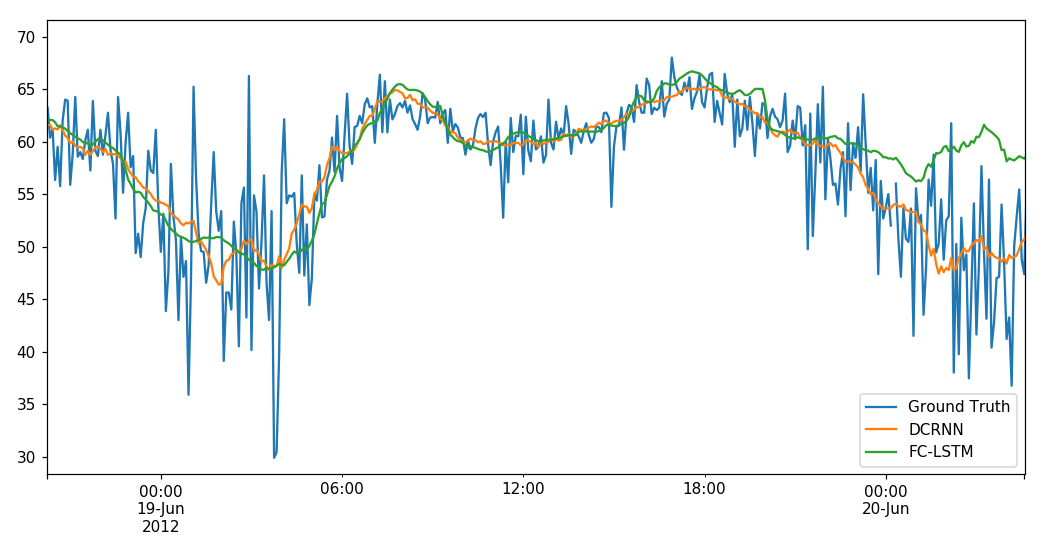}
    \includegraphics[width=0.8\linewidth]{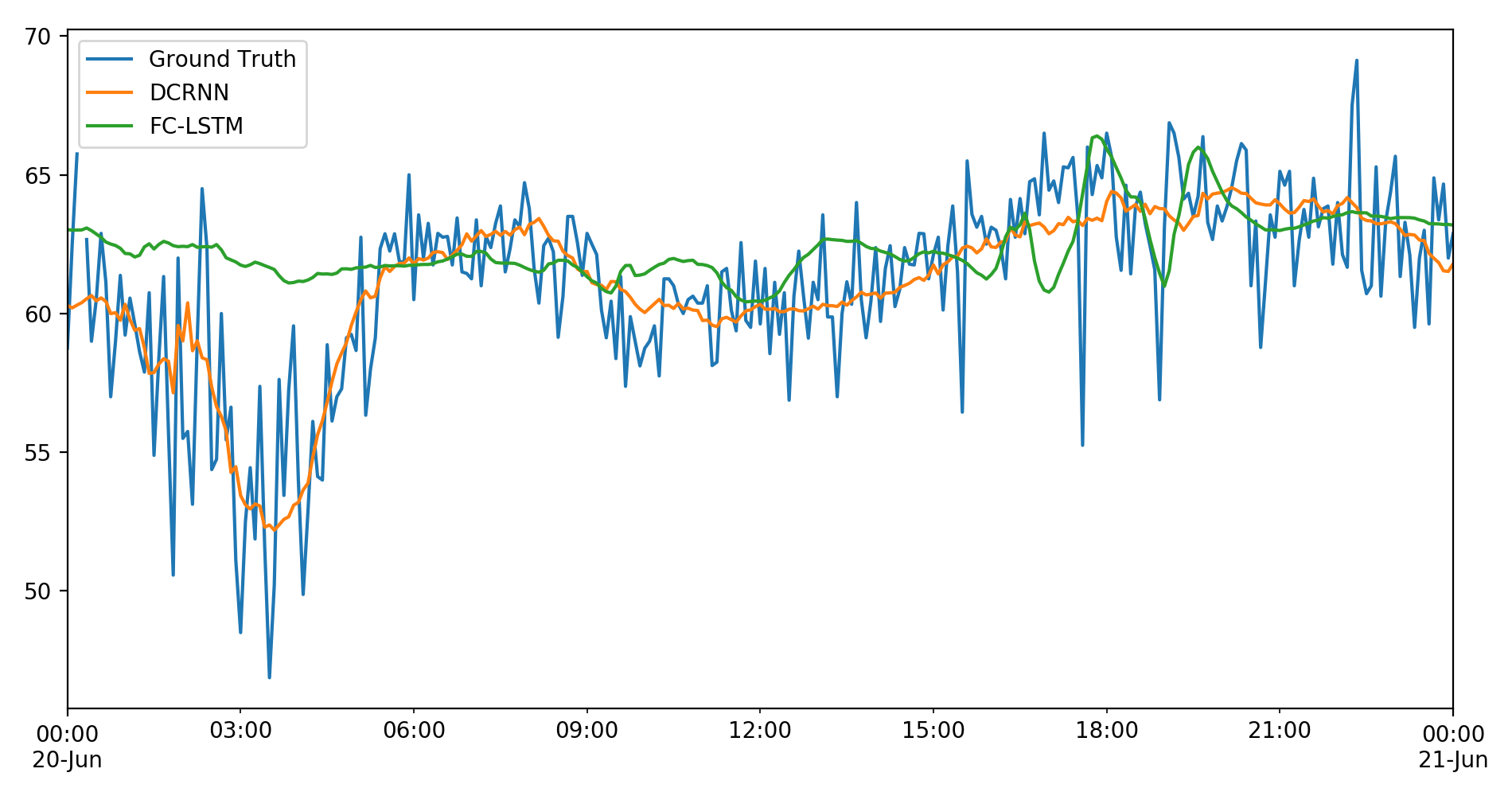}
	\caption{Traffic time series forecasting visualization.}
	\label{fig:traffic_visualization_large3}
\end{figure*}

\end{document}